\documentclass[runningheads]{llncs}

\usepackage{eccv}

\usepackage{eccvabbrv}

\usepackage{graphicx}
\usepackage{booktabs}
\usepackage{array}
\newcolumntype{P}[1]{>{\centering\arraybackslash}p{#1}}
\usepackage{tikz}
\usepackage{comment}
\usepackage{amsmath,amssymb} 
\usepackage{bbm}
\usepackage{color}
\usepackage{colortbl}
\usepackage{multirow}
\usepackage{graphicx}
\usepackage{subcaption}
\usepackage[table]{xcolor}
\definecolor{lightgray}{gray}{0.93}
\usepackage{pifont}
\usepackage{adjustbox}
\usepackage{multirow}
\usepackage{tikz}

\usepackage{wrapfig}
\usepackage{soul}

\definecolor{darkred}{RGB}{150,0,0}

\newcommand{\App}{\textbf{\textcolor{darkred}{Appendix}}} 

\newcommand{\ours}{QaTS}

\definecolor{lightgray}{gray}{0.92}

\newcommand{\vp}{\mathbf{p}}
\newcommand{\vl}{\mathbf{l}}

\newcommand{\calX}{\mathcal{X}}

\newcommand{\calD}{\mathcal{D}}
\newcommand{\calY}{\mathcal{Y}}

\newcommand{\real}{\mathbb{R}}

\definecolor{brightgray}{RGB}{220,220,220}
\definecolor{lightgraycol}{gray}{0.92}

\usepackage{pifont}

\newrobustcmd*{\mytriangle}[1]{\tikz{\filldraw[draw=#1,fill=#1] (0,0) --
(0.1cm,0) -- (0.05cm,0.1cm);}}

\usepackage{makecell}

\usepackage[accsupp]{axessibility}  
\usepackage{pifont}

\usepackage{array}

\usepackage{colortbl} \definecolor{oodcol}{RGB}{245,230,230} 

\newcolumntype{P}[1]{>{\centering\arraybackslash}p{#1}}

\usepackage[table]{xcolor} 
\usepackage{color} 

\definecolor{Better}{rgb}{0.18, 0.407, 0.266}
\definecolor{Worse}{rgb}{0.35, 0.35, 0.35}

\definecolor{ours_color}{gray}{0.9}

\newcommand{\imp}[1]{$_{{\textbf{\textcolor{Better}{#1}}}}$}

\usepackage{hyperref}

\usepackage{orcidlink}

\begin{document}

\title{Quantile‑Adaptive Temperature Scaling for Confidence Calibration} 




\titlerunning{Abbreviated paper title}

\author{Omprakash Chakraborty\inst{1}\orcidlink{0000-0001-6050-8031} \and
Leo Fillioux\inst{2}
\and
Ismail Ben Ayed\inst{1}\orcidlink{0000-0002-9668-8027}
\and
Jose Dolz\inst{1}\orcidlink{0000-0002-2436-7750}}

\authorrunning{O.~Chakraborty et al.}

\institute{ÉTS Montréal, Canada \and
Université Paris-Saclay, CentraleSupélec, Gustave Roussy, INSERM, CDSU, IHU PRISM, Gif-sur-Yvette}

\maketitle

\begin{abstract}
Deep neural networks often produce poorly calibrated confidence estimates, overstating their certainty even when predictions are incorrect. Temperature Scaling (TS) remains the most widely used post‑hoc calibration method due to its simplicity and effectiveness, yet its global, uniform rescaling of logits fails to correct the highly heterogeneous structure of miscalibration observed across the confidence spectrum. In particular, the largest correctness–confidence discrepancies arise in different quantile regions depending on the setting, 
which standard TS leaves largely unaddressed. We introduce Quantile‑Adaptive Temperature Scaling (\ours{}), a simple and efficient post‑hoc calibration method that adapts the temperature as a function of a prediction’s empirical confidence quantile. By mapping confidences into the quantile space, \ours{} normalizes the calibration problem, makes the structure of miscalibration explicit, and enables a monotone temperature function that adapts  across quantiles while leaving well‑calibrated high‑confidence predictions largely unchanged. 
This quantile‑aware formulation aligns naturally with a reparameterized Expected Calibration Error (ECE) objective and yields a sample‑wise temperature that is robust across a variety of challenging scenarios, such as class imbalance and distributional shifts. Across a broad range of datasets, architectures, evaluation scenarios and diverse tasks, 
\ours{} consistently, and substantially, outperforms state‑of‑the‑art post‑hoc calibration methods, delivering more reliable and trustworthy confidence estimates without modifying model predictions. Our code is available at \href{https://github.com/omchakrabarty/QaTS}{\texttt{https://github.com/omchakrabarty/QaTS}}.
\end{abstract}

\section{Introduction}
\label{sec:introduction}


Confidence calibration, the degree to which predicted probabilities reflect actual correctness, plays a vital role in ensuring reliable risk estimation for decision‑making systems. Poor calibration can distort perceived failure risk and, in turn, drive suboptimal or misguided decisions. Although deep learning models have substantially improved predictive discrimination, they often suffer from over‑confidence, a widely recognized shortcoming that makes calibration a persistent challenge \cite{guo2017calibration,minderer2021revisiting}. This issue becomes especially consequential in high‑stakes domains such as healthcare, autonomous driving, and critical infrastructure, motivating a substantial body of work on improving calibration of deep models \cite{larrazabal2023maximum,murugesan2023trust,liu2022devil,mukhoti2020calibrating}.


While training-based \cite{liu2022devil,liu2023class,mukhoti2020calibrating} and post-hoc \cite{guo2017calibration,hekler2023test,tomani2022parameterized} confidence calibration approaches exist, the latter offer a more appealing and practical solution in real-world settings, since they avoid costly retraining, do not require access to model weights or gradients, and scale seamlessly to large pretrained models and frozen backbones. In particular, Temperature scaling (TS) \cite{guo2017calibration}, a single- parameter variant of Platt Scaling \cite{platt1999probabilistic}, calibrates a model by applying a single scalar temperature to all logits, uniformly shrinking or expanding confidence scores without altering the prediction. 
Several extensions have been proposed, including class‑wise \cite{jung2023scaling}, prediction-wise \cite{tomani2022parameterized} or input‑dependent \cite{ding2021local} temperatures, often learned through small auxiliary networks. While these variants increase flexibility, they all share the same underlying assumption: miscalibration can be corrected by a smooth, globally consistent rescaling function that applies uniformly across the confidence spectrum. 

\begin{figure*}[!h]
    \centering
    \includegraphics[width=\textwidth]{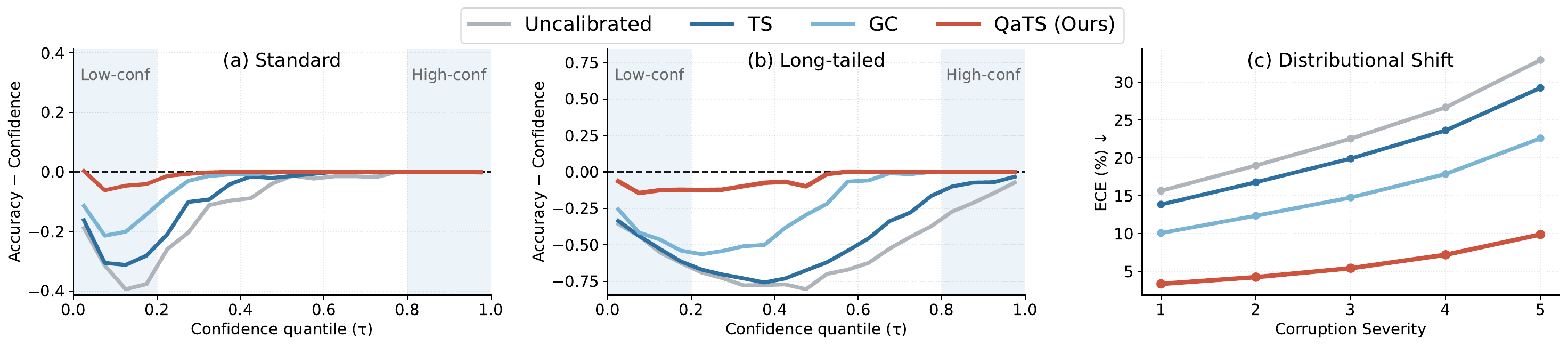}
    \caption{
    \textbf{Calibration behavior across regimes.}
    Calibration gap ($\mathrm{Acc} - \mathrm{Conf}$) under (a) \emph{standard} and (b)  \emph{long-tailed} setting.
    (c) ECE under increasing severity of \emph{distributional shift}.
    Across all regimes, \ours{} consistently reduces calibration bias and error, while maintaining robustness under severe distributional shifts.
    }
    \label{fig:teaser}
\end{figure*}

Nevertheless, our quantile-wise analysis in Fig. \ref{fig:teaser} reveals that miscalibration (\textit{gray}) is far from uniform across the whole confidence spectrum. In the standard regime (\textit{left plot}), the largest correctness–confidence gaps consistently occur in the lowest confidence quantiles, precisely where uncertainty is highest and reliable confidence matters most, while mid‑ and high‑quantile predictions exhibit substantially smaller discrepancies. In the long-tailed regime (\textit{middle}), this structure changes: the dominant errors shift toward the medium-confidence quantiles, driven by the prevalence of mid-frequency classes that produce many moderately confident yet error-prone predictions. Across these settings, both the shape and location of miscalibration vary substantially, indicating that calibration errors are inherently heterogeneous and regime-dependent. This pronounced heterogeneity further exposes a structural limitation of existing temperature‑scaling approaches, i.e., they implicitly assume that miscalibration can be corrected by a single, smooth, globally consistent transformation of the logits. Such uniformity assumptions are fundamentally mismatched with the empirical behavior shown in Fig.~\ref{fig:teaser}. Methods like TS \cite{guo2017calibration} and its more advanced versions, such as GC \cite{yang2023beyond}, 
apply a single functional mapping from logits to temperatures, treating all confidence levels as if they require the same type of correction. Because they do not condition on confidence quantiles, they cannot effectively adapt their corrections to regions where miscalibration is more severe. 
As a result, these methods often reduce global ECE while leaving the dominant 
quantile-wise discrepancies substantially uncorrected. 

A direct implication of this structural rigidity is the degraded performance of TS-based methods under more challenging scenarios, such as distributional drifts (Fig.~\ref{fig:teaser}, \textit{right}). Covariate perturbations often alter the absolute scale of confidence values, while largely preserving their relative ordering. 
Because the temperature function $T(x)$ in \ours{} depends on the empirical confidence quantile $\tau$ rather than the raw confidence magnitude $p$ (as in TS-based approaches), it inherits the rank-based stability of quantiles, i.e., 
they are stable under monotone transformations of the logits, and thus substantially less sensitive to global logit scaling effects that often accompany covariate shift, as long as the relative ordering of confidences is approximately preserved. As a result, \ours{} applies its calibration correction consistently across quantiles, even when test samples differ from calibration samples. As long as the ordering of confidence values is approximately preserved (a mild assumption satisfied by most covariate shifts) the quantile-adaptive temperature continues to reduce the dominant contributions to the calibration error in test samples, yielding robust calibration across domains. 
These observations motivate a calibration strategy that operates directly in quantile space, where the structure of miscalibration becomes explicit and can be corrected in a targeted, distribution‑aware manner.

To address these limitations, we propose Quantile‑Adaptive Temperature Scaling (\ours{}), a calibration method explicitly designed to correct the heterogeneous structure of miscalibration revealed by our quantile‑wise analysis. Instead of applying a single global temperature, or predicting temperatures from logits, features, or auxiliary networks, \ours\;introduces a sample‑wise temperature $T(x)$ that adapts to the empirical confidence quantile of each prediction. By mapping raw confidences into quantile space, \ours\;obtains a normalized, distribution‑aware measure of how “typical” or “atypical” a prediction is relative to the calibration set. This enables a monotone temperature function that selectively softens low‑quantile (overconfident) predictions and sharpens high‑quantile ones, improving calibration without altering prediction rankings. Combined with surrogate‑based training using strictly proper scoring rules, \ours{} provides a simple, efficient, and robust mechanism for correcting miscalibration exactly where it is most severe, yielding confidence estimates that remain reliable even under distribution shift.



We can summarize our contributions as follows:
\begin{itemize}
\item We provide a quantile-wise analysis of miscalibration, revealing that calibration errors are highly heterogeneous across the confidence quantile spectrum, and vary substantially across scenarios (e.g., standard vs long-tailed). This analysis reveals structural limitations of existing TS‑based methods, which cannot directly target the regions where miscalibration is most pronounced.


\item Based on these observations, we introduce a simple yet effective strategy, \ours{}, for calibrating neural networks that directly targets the heterogeneous structure of miscalibration, operating directly in quantile space. We formalize a quantile reparameterization of the Expected Calibration Error (ECE), which expresses calibration error as a function of confidence quantiles rather than raw confidence values. This reparametrization exposes where miscalibration is concentrated and enables a temperature function that adapts to the relative rank of each prediction, yielding calibration that is quantile‑aware, showing practical benefits in challenging scenarios. 

\item Comprehensive experiments across multiple benchmarks and distinct tasks, including standard computer vision and medical image classification, text classification and semantic segmentation, as well as diverse scenarios, such as long-tailed distributions and distributional drifts, show that \ours{} yields superior uncertainty calibration than state-of-the-art post-hoc methods.

\end{itemize}

\section{Related Work}
\label{sec:related_work}

\noindent \textbf{Uncertainty calibration} of deep neural networks has emerged in the last years as an active area of study, with numerous studies analyzing and characterizing the calibration behavior of modern models \cite{guo2017calibration,minderer2021revisiting}. These methods can be mainly categorized into: \textit{training-time} and \textit{post-hoc} calibration. 

\noindent \textit{Training-time} calibration aims at enhancing model calibration during training, typically by integrating additional 
regularization objectives into the loss function \cite{pereyra2017regularizing,cheng2022calibrating,park2023acls,murugesan2023trust,liu2022devil,liu2023class,jung2023scaling}. 
A popular strategy is to introduce explicit penalties that either penalize overconfident softmax predictions \cite{pereyra2017regularizing,larrazabal2021maximum,cheng2022calibrating,park2023acls} or encourage small logit differences~\cite{murugesan2023trust,liu2022devil,liu2023class}. Beyond such explicit objectives, several works have shown that widely used classification losses, such as Focal Loss \cite{lin2017focal} and Label Smoothing \cite{szegedy2016rethinking}, implicitly promote higher‑entropy predictions, thereby yielding models with lower confidence \cite{muller2019does,mukhoti2020calibrating}. Additional training‑time techniques 
include data‑mixing strategies such as MixUp \cite{thulasidasan2019mixup,zhang2022and} and constraints on the geometry of the logit space, for example by enforcing a fixed logit‑norm \cite{wei2022mitigating} or logit-range \cite{murugesan2024robust}. A common limitation of training‑time methods is that they are tightly coupled to the optimization of the underlying classifier, which restricts their use to settings where the model can be modified and retrained. 

\noindent \textit{Post-hoc calibration}, in contrast, adjusts the output logits of a pre‑trained model 
without modifying its training procedure  \cite{tomani2022parameterized,joy2023sample,guo2017calibration,ovadia2019can,hekler2023test,guptacalibration}, offering the practical advantage that calibration can be applied independently of the training process and even to fixed or proprietary models. A notable method is temperature scaling (TS) \cite{guo2017calibration}, a simplified 
form of Platt scaling \cite{platt1999probabilistic}, which applies a single global temperature $T$. 
While effective at reducing overconfidence on average, 
global scaling cannot adapt to input‑dependent uncertainty, and may under‑ or over‑correct different confidence ranges. More flexible non‑parametric techniques, such as Histogram Binning \cite{niculescu2005obtaining} and Isotonic Regression (IR) \cite{zadrozny2002transforming}, learn piecewise mappings that can substantially improve calibration. Nevertheless, they lack structural constraints such as smoothness or monotonicity beyond local segments, which can lead to unstable mappings that generalize poorly to unseen confidence ranges. Recent work explores parameterized temperature scaling, where temperatures depend on the class \cite{frenkel2021network}, or on the input via a small auxiliary network \cite{tomani2022parameterized,ding2021local}, 
increasing flexibility while retaining the accuracy‑preserving nature of classical scaling. Furthermore, improving calibration has recently been explored from the perspective of the feature space, either by modulating the temperature as a function of learned features \cite{yang2023beyond,xiong2023proximity} or by selectively clipping feature magnitudes to increase predictive entropy on high‑error samples while preserving discriminative information on well‑calibrated ones \cite{tao2025feature}. 
AdaTS \cite{hekler2023test} leverages test‑time augmentation by generating multiple transformed versions of each input and averaging their logits to obtain a better‑calibrated prediction, but this comes at the cost of substantially increased inference‑time computation. 
Other approaches have extended TS to multiple domains \cite{yu2022robust}, or under distributional drifts \cite{yu2022robust,kim2024uncertainty,tomani2021post}, but they require either direct access to 
samples of each domain \cite{yu2022robust,kim2024uncertainty,gong2021confidence}, or synthetically generate larger validation sets through image transformations \cite{tomani2021post}, which attempt to simulate domain shift conditions.









In contrast to all existing techniques, the proposed \ours\; relies on confidence quantiles, rather than in raw confidence scores, which makes the method inherently robust to the scale, distribution, and calibration quality of the underlying classifier. By operating on rank‑based information, \ours\; preserves ordering while eliminating dependence on absolute score values, enabling consistent behavior across architectures, datasets, diverse tasks and scenarios. 

\section{Problem Definition and Motivation}
\label{sec:problem_definition}

\noindent \textbf{Preliminaries.} Let $\calX\subset\mathbb{R}^d$ denote the input space and $\calY=\{1,\dots,K\}$ the label space. We assume a $K$-class trained classifier is a mapping function $\phi:\calX\rightarrow\Delta_K,$ where $\Delta_K=\{\vp \in\mathbb{R}^K:\sum_{k=1}^K p_k=1,\; p_k\ge 0\}$ is the probability simplex. Furthermore, the probability predictor $\phi$ is composed of a feature extractor $g:\calX\rightarrow\mathbb{R}^K$ that produces the logits vector $\vl=(\ell_{k})_{1 \leq k \leq K}$, followed by the softmax operator, defined by $\texttt{sm}$, so that $\phi=\boldsymbol{\texttt{sm}} \circ g $. Thus, for an input $x\in\calX$, 
we can compute the 
softmax probability vector $\vp(x)=\phi(x)=(p_1(x),\dots,p_K(x))$, as
\begin{equation}
p_k(x) = \frac{\exp(\ell_k(x))}{\sum_{j=1}^K \exp(\ell_j(x))}, \quad k=1,\dots,K.
\end{equation}
From this vector, we can now compute the predicted class as $\hat{y}(x) = \arg\max_k p_k(x)$, whose associated confidence is $\text{conf}(x) = \max_k p_k(x)$. 

In our post-hoc calibration scenario, we have access to a pre-trained model $\phi$, and a calibration set $\calD_{cal}$ of $\calX \times \calY$, where $X \in \calX$ contains \textit{i.i.d.} samples from an unknown distribution with $Y \in \calY$. We can now define a post-hoc calibration function $f : \real^K \to \real^K$, yielding the K-class \textit{calibrated} predictor as $\phi_c=\boldsymbol{\texttt{sm}} \circ f \circ g.$

\begin{definition}[Perfect Calibration]
\label{def1}
Let $(X,Y)$ be random variables taking values in 
$\calX\subset\mathbb{R}^d$ and $\calY=\{1,\dots,K\}$,  respectively, drawn from an unknown distribution $\pi$.
Let $\phi_c:\calX\to\Delta_K$ be a $K$-class probability predictor, with the definitions of $\hat{y}(X):=\arg\max_k p_k(X)$, and $\hat{p}(X):=\max_{k} p_k(X)$, where $\vp (X)= \phi_c(X),$ we say that $\phi_c$ is perfectly calibrated if:
\begin{equation}
   \mathbb{P}\big(Y=\hat{y}(X)\mid \hat{p}(X)=p\big)=p,
\qquad \forall\, p\in[0,1]. 
\end{equation}

\end{definition}

Thus, our objective is to train a post-hoc calibration function $f$ that satisfies \textit{Definition \ref{def1}}, 
i.e., yields a perfectly calibrated predictor. To this end, we optimize the Negative Log Likelihood (NLL) loss, a standard choice in calibration \cite{guo2017calibration,zhang2020mix}.

\noindent \textbf{Expected calibration error as a discrepancy between correctness and confidence.} Following Definition~\ref{def1}, let $C(p)$ denote the \textit{correctness} function for a scalar confidence value $p\in[0,1]$ as 
\begin{equation}
    C(p) = \mathbb{P}(Y=\hat{y}(X)\mid \text{conf}(X)=p), 
\end{equation}
and let $\mu$ denote the distribution of $\text{conf}(X)$ on [0,1], where $\text{conf}(X):=\hat{p}(X)$. The Expected Calibration Error (ECE), in expectation, can be then written as:
\begin{equation}
\mathrm{ECE} = \int_0^1 \big| C(p) - p \big| \, d\mu(p),
\label{eq:ece_integral}
\end{equation}
which measures the $L^1 (\mu)$ distance between correctness and confidence.

\noindent \textbf{Quantile reparameterization.} The metric in Eq. \ref{eq:ece_integral} weights miscalibration in a pointwise manner, based on how often each confidence value occurs. In this scenario, dense regions dominate on the final value, whereas rare confidence values contribute very little. This leads to large errors concentrated in rare confidence regions to be masked by small errors in common regions, yielding a misleading low global ECE. To avoid this, we first define an empirical quantile function. 
To do so, we first sort the confidences of all calibration samples, $\text{conf}(x^{(1)}) \leq \text{conf}(x^{(2)}) \leq \dots \leq \text{conf} (x^{(N)})$, and define the empirical quantile function as: 
\begin{equation}
F_{\text{conf}}(u) = \frac{1}{N} \sum_{i=1}^N \mathbbm{1} {\{\text{conf}(x^{(i)})\le u\}},
\end{equation}
which returns the empirical cumulative distribution function (CDF) value of a given input confidence $u$. Thus, for an image $x$, we can first compute its 
empirical quantile as:
\begin{equation}
\label{eq:emp_quant}
q(x) = F_{\text{conf}}(\text{conf}(x)).
\end{equation}

Now, we can reparameterize Eq. \ref{eq:ece_integral} by quantiles $\tau=F_{conf}(p)$, so that the same error is integrated uniformly over $\tau \in [0,1]$. 
To do this, let $q(x) = F_{\mathrm{conf}}(\text{conf}(x))$ be the empirical quantile of the confidence. Under mild regularity assumptions, $q(X)$ is asymptotically uniform on $[0,1]$ and the change of variables $p = F_{\mathrm{conf}}^{-1}(\tau)$ yields the equivalent representation
\begin{equation}
\mathrm{ECE} 
= \int_0^1 
\big| C(F_{\mathrm{conf}}^{-1}(\tau)) - F_{\mathrm{conf}}^{-1}(\tau) \big| \, d\tau.
\label{eq:ece_quantile}
\end{equation}
This reparameterization leaves ECE unchanged but expresses it on a uniform quantile axis, preventing dense confidence regions from dominating the representation of calibration error. Under this view, 
ECE becomes the $L^1$ discrepancy between correctness and confidence \emph{expressed along the uniform quantile axis}. This representation highlights that miscalibration is often \emph{heterogeneous} across quantiles, and can vary under different conditions, as empirically shown in Fig.~\ref{fig:teaser}. 


\section{Adaptive Quantile Temperature Scaling}
\label{sec:methodology}


\noindent \textbf{Sample-dependent Temperature.} To account for individual sample uncertainties when scaling the logits in the softmax function, we propose a sample-wise temperature function, 
defined as a monotone linear function of $(1-q(x))$:
\begin{equation}
\label{eq:temp_ours}
    T(x) = a \cdot (1 - q(x)) + b, \quad q(x) \in [ 0,1],
\end{equation}
where $a$ and $b$ are two positive learnable parameters that modulate the final sample temperature value according to its quantile.  
This ensures that 
lower‑quantile (i.e., more problematic) predictions receive a different scaling than higher‑quantile (i.e., less problematic) ones. 
Unlike standard TS-based approaches, \ours{} conditions the temperature on the sample’s confidence quantile, not on its raw logit magnitude. This quantile‑based parameterization breaks the global‑smoothness assumption and enables corrections that vary systematically across the confidence spectrum, aligning the calibration mechanism with the heterogeneous structure revealed by the quantile‑wise analysis (Fig.~\ref{fig:teaser}).

\noindent \textbf{Learning the temperature parameters $a$ and $b$.} The parameters $a$ and $b$ are optimized using the calibration (or validation) set so that the scaled probabilities
\begin{equation}
\tilde{p}_k(x) = \frac{\exp(\ell_k(x)/T(x))}{\sum_{j=1}^K \exp(\ell_j(x)/T(x))}    
\end{equation}
better reflect empirical correctness frequencies.

Direct optimization of Expected Calibration Error (ECE) is infeasible due to its non-differentiable binned form. Instead, we fit the parameters $(a,b)$ using smooth surrogates such as the negative log-likelihood (NLL), 
a strictly proper scoring rule. Minimizing this surrogate encourages the calibrated probabilities to align with empirical correctness frequencies, thus reducing calibration error in practice. Specifically, we minimize 
the 
NLL over the set of calibration samples: 
\begin{equation}
\min_{a,b>0} \; \frac{1}{N} \sum_{i=1}^{N} 
\mathcal{L}\big(\tilde{\mathbf{p}}(x^{(i)}), \mathbf{y}^{(i)}\big)
= 
\min_{a,b>0} \; -\frac{1}{N} \sum_{i=1}^{N} \sum_{k=1}^K y^{(i)}_k \log (\tilde{p}_k(x^{(i)};a,b)).
\end{equation}
where $\mathbf{y}^{(i)}$ is the ground-truth label of calibration sample $x^{(i)}$, and $\mathcal{L}$ denotes the chosen loss function. 
We also constraint $a,b>0$, ensuring a strictly positive, monotonically decreasing temperature over $q \in [0,1]$. 
This training procedure yields optimal values of $a^*$ and $b^*$ that adaptively adjust the temperature based on each prediction’s confidence quantile, directly targeting the heterogeneity revealed by the quantile reparameterization and thereby improving ECE. 

\noindent \textbf{Inference.} For each test sample $x$, we simply scale each predicted logit by the sample-wise temperature scale function in \cref{eq:temp_ours}, which dynamically adapts to the underlying properties of each sample:
\begin{equation}
\label{eq:logits}
\tilde{\ell}_k(x) = \frac{\ell_k(x)}{T(x)}, \quad k=1,\dots,K.
\end{equation}

\noindent \textbf{Benefits of the proposed approach.} Considering the definition of the empirical quantile in Eq. \ref{eq:emp_quant}, our method introduces a quantile-dependent temperature value, defined in Eq. \ref{eq:temp_ours}, which is used to rescale the logits in (\ref{eq:logits}). As $q(x)$ indexes the position of a prediction along the uniform quantile 
axis, adjusting $T(x)$ as a function of $q(x)$ directly modifies the confidence 
\emph{precisely in the regions where Eq.~\eqref{eq:ece_quantile} shows the 
largest contribution to ECE}. In particular, \textit{i)} low-quantile predictions (where overconfidence is typically largest) receive higher temperatures, reducing $p$ and decreasing $|C(p)-p|$ in Eq.~\eqref{eq:ece_quantile}; and \textit{ii)} high-quantile predictions receive smaller adjustments, preserving accuracy and avoiding unnecessary distortion. Since the quantile axis is uniform, any reduction of the discrepancy in the 
regions with the largest quantile-wise error leads to a monotone decrease of 
the global ECE. Thus, quantile-adaptive temperature scaling is explicitly aligned with minimizing Eq.~\eqref{eq:ece_quantile}. 
Beyond this alignment with the calibration objective, the quantile representation also provides a key advantage under challenging scenarios, such as domain shift. 
Mapping confidences through the calibration CDF yields a monotone transform, so \ours{} depends mainly on the relative ordering of scores rather than their absolute scale. This makes the temperature function more stable under shifts that preserve approximate ranking, even when logits are globally rescaled or shifted. By learning a simple temperature function $T(x) = a \cdot (1 - q(x)) + b$ on these quantile values, the model captures how calibration should vary as a function of relative confidence, not absolute scale. This means that, at deployment, even in the presence of distributional drifts, the ordering of predictions remains far more consistent than their raw magnitudes on these quantile values, the model captures how calibration should vary as a function of \emph{relative} confidence rather than absolute scale. 

\section{Experiments}
\label{sec:experiments}

\noindent \textbf{Datasets.} \ours\;is empirically validated across several computer vision classification tasks, including {CIFAR-10} and {CIFAR-100}~\cite{krizhevsky2009cifar}, {ImageNet}~\cite{russakovsky2015imagenet}, as well as {their long-tailed counterparts}~\cite{cao2019learning}, with different degrees of data imbalance ratios $\gamma$, 
and {ImageNet-LT}~\cite{liu2019imagenetlt}. 
Further, we evaluate the robustness of \ours{} under distributional shifts by employing corrupted benchmarks such as {CIFAR-100-C}~\cite{hendrycks2019robustness}, which introduces 15 corruption types 
across 5 severity levels. To assess the performance of the proposed method in real-world scenarios, we perform experiments in medical imaging classification tasks, by resorting to {HAM10000}~\cite{tschandl2018ham10000} and to MedMNIST2D \cite{medmnistv1,medmnistv2}, which consists of 10 pre-processed 2D sources covering primary data modalities (e.g., X-Ray, OCT, Ultrasound, and CT), and diverse binary and multi-class classification tasks. 
Last, to show the general applicability of our approach, we further evaluate it on the {20 Newsgroups} dataset~\cite{twenty_newsgroups_113}, a popular Natural Language Processing benchmark for text classification and PascalVOC~\cite{everingham2015pascal}, a well-established image segmentation dataset. Thus, we evaluated \ours{} across \textbf{162 different settings.} 

\noindent \textbf{Baselines.} \textbf{\textit{Accuracy preserving methods:}} We benchmark \ours{} to post-hoc calibration approaches that preserve accuracy, including the raw uncalibrated model (Uncal), temperature scaling (TS)~\cite{guo2017calibration}, Isotonic Regression (IR)~\cite{zadrozny2002transforming}, and the more recent AdaTS \cite{hekler2023test} and Group Calibration (GC) \cite{yang2023beyond}, where we report the values of the proposed method combined with temperature scaling, which yields the most competitive performance. \textbf{\textit{Non-accuracy preserving methods:}} We also include FeatClip \cite{tao2025feature}, 
a very recent post-hoc calibration approach that modifies model predictions. Importantly, as highlighted in prior works \cite{yang2023beyond}, such approaches often have a noticeable negative impact on model accuracy. Our empirical evaluation confirms this trend, where FeatClip's accuracy degradation becomes substantial in challenging settings such as class‑imbalanced scenarios. 

\noindent \textbf{Evaluation metrics.} To measure the discriminative performance of the different methods, we use classification accuracy (ACC) for classification tasks, and mean Intersection over Union (mIoU) for segmentation. In terms of calibration, we follow the standard literature and resort to the Expected Calibration Error (ECE). In particular, with $N$ samples grouped into $M$ bins $\{b_1, b_2, \ldots, b_{M}\}$, the ECE is calculated as: $\sum_{m=1}^{M} \frac{\left|b_{m}\right|}{N}\left|\operatorname{acc}\left(b_m\right)-\operatorname{conf}\left(b_m\right)\right|$, where $\operatorname{acc}\left(\cdot\right)$ and $\operatorname{conf}\left(\cdot\right)$ denote the average accuracy and confidence in bin $b_{m}$. 

\noindent \textbf{Implementation Details.} For image classification, we experiment with multiple convolutional neural networks, such as ResNet-18, ResNet-50 and DenseNet-121, as well as Vision Transformers, including ViT-S/16 and ViT-B/16. We use $M = 15$ bins for ECE computation across all datasets and architectures. Furthermore, we use BERT-B~\cite{devlin2019bert} on the NLP recognition task and DeepLabV3~\cite{chen2017rethinking} 
for semantic segmentation on PASCAL VOC2012. 

\subsection{Results}
\begin{figure}[!t]
    \centering
    \includegraphics[width=0.9\linewidth]{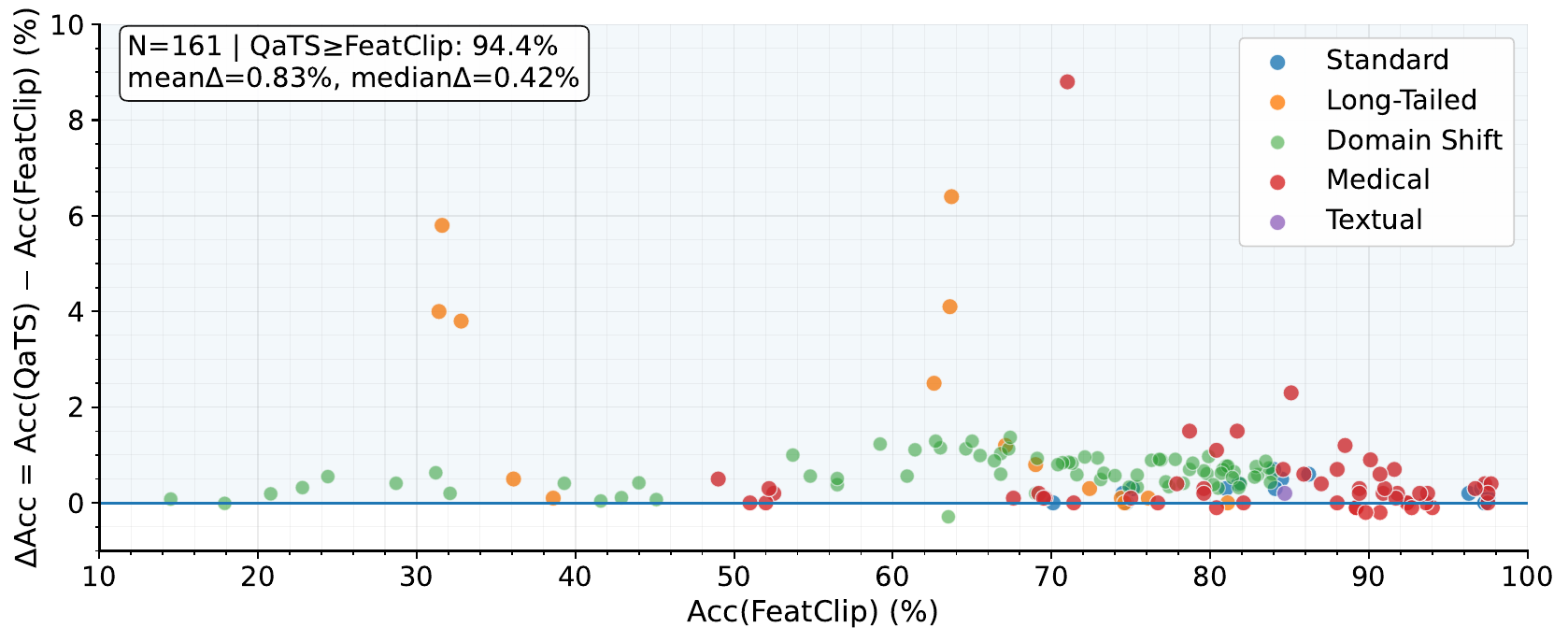}
    \caption{
    \textbf{Accuracy dominance of \ours{} over FeatClip across 161 evaluation settings spanning standard, long-tailed, domain-shift, medical and text classification benchmarks.} Each point corresponds to the Top-1 accuracy difference for a dataset-architecture pair. Points above zero indicate better accuracy for \ours{}. 
    }
    \label{fig:accuracy}
\end{figure}

\noindent \textbf{A note on classification performance.} Post-hoc calibration approaches are often designed to operate by reshaping the softmax (or logit) distributions, without altering the predicted class. Nevertheless, the underlying mechanism of several recent methods improve calibration, but at the cost of degrading their discriminative performance. 
FeatClip~\cite{tao2025feature}, a recent state-of-the-art post-hoc calibration approach, is a notable example: \textit{its calibration gains come with measurable accuracy degradation.} To contextualize the calibration results presented in this section, we first report the classification accuracy of FeatClip and compare it to our method, which preserves the predicted class by design. This allows readers to appreciate how certain calibration techniques may degrade accuracy before focusing solely on their calibration behavior. 
In particular, Fig. \ref{fig:accuracy} presents 
a pair-wise comparison across 161 dataset–architecture combinations spanning standard, long-tailed, domain-shift, medical and text classification benchmarks. Each point corresponds to the accuracy difference between \ours{} and FeatClip ($\Delta$Acc). 
Overall, \textbf{\ours\;yields higher 
accuracy in~94\% of the settings.} 
Importantly, these gains are consistent across diverse regimes, indicating that the calibration improvements of \ours\;are not obtained at the expense of discriminative performance. This reinforces the importance of assessing calibration methods alongside classification accuracy rather than in isolation.

\begin{table}[!h]
\centering
\scriptsize
\caption{\textbf{Calibration performance (ECE$\downarrow$) in standard image classification benchmarks.} Results across datasets and backbones, where $^\ddagger$ indicates that the method modifies model predictions, which impacts the final accuracy. Best method in \textbf{bold}, whereas second best \underline{underlined}. Performance gap wrt second best method is highlighted as \textbf{\textcolor{Better}{better}} or \textbf{\textcolor{Worse}{worse}}. Same convention applies throughout the paper.}
\label{tab:main_clean_ece}
\begin{tabular}
{p{1.6cm}lP{1cm}P{1.cm}P{1.cm}P{1.cm}P{1.cm}P{1.3cm}P{1.3cm}}

\toprule
\textbf{Dataset} & \textbf{Backbone} & \textbf{Uncal} & \textbf{TS}& \textbf{IR} &  \textbf{AdaTS} & \textbf{GC} &\textbf{FeatClip}$^\ddagger$ & \textbf{\ours} \\

\midrule

\multirow{5}{*}{CIFAR-10}
 & RN-50  & 1.71 & 1.23 & 0.64 &  0.64 & 0.78 &\underline{0.55} & \textbf{0.31}\imp{-0.24} \\
 & RN-18  & 1.93 & 0.77 & 0.59 & 0.62 &  0.70 &\underline{0.50} & \textbf{0.32}\imp{-0.18} \\
 & DN-121 & 1.44 & 1.02 & 0.67 & 0.62 &  \underline{0.50} & 0.58 & \textbf{0.29}\imp{-0.21} \\
 & ViT-S  & 1.80 & 1.01 & \textbf{0.45} &  0.67 & 1.01 & 0.62 & $\underline{0.51}_{+0.06}$ \\
 & ViT-B  & 1.97 & 1.61 & 0.65 & 1.18 & 1.19 & \underline{0.54} & \textbf{0.49}\imp{-0.05} \\
\midrule

\multirow{5}{*}{CIFAR-100}
 & RN-50  & 8.62 & 5.32 & 3.08 &  3.93 & 3.99 &\underline{3.01} & \textbf{2.56}\imp{-0.45} \\
 & RN-18  & 7.87 & 3.18 & 1.71 &  2.91 & 2.37 &\underline{1.70} & \textbf{1.32}\imp{-0.38} \\
 & DN-121 & 7.60 & 5.15 & 2.91 & 3.12 & 3.12 & \underline{2.71} & \textbf{2.45}\imp{-0.26} \\
 & ViT-S  & 7.98 & 6.74 & \underline{2.73} &  3.62 & 3.37 & 2.92 & \textbf{2.54}\imp{-0.19} \\
 & ViT-B  & 10.79 & 9.43 & 4.67 & 5.34 & 5.35 & \underline{2.57} & \textbf{2.26}\imp{-0.31} \\
\midrule

\multirow{5}{*}{ImageNet}
 & RN-50  & 3.71 & 2.26 & 2.72 &  2.05 & 2.21 & \underline{1.67} & \textbf{1.58}\imp{-0.09} \\
 & RN-18  & 2.61 & 1.86 & 1.50 &  1.71 & 1.84 & \underline{1.41} & \textbf{1.34}\imp{-0.07} \\
 & DN-121 & 2.61 & 1.89 & 2.10 &  1.84 & \underline{1.79} & 1.81 & \textbf{1.71}\imp{-0.08} \\
 & ViT-S  & 2.11 & 1.95 & 2.03 & \underline{1.28} &  1.51 & 1.51 & \textbf{1.13}\imp{-0.15} \\
 & ViT-B  & 5.77 & 3.28 & 2.80 & 3.01 & 3.28 &  \underline{2.93} & \textbf{2.49}\imp{-0.44} \\
\bottomrule
\end{tabular}%
\end{table}

\noindent \textbf{A. Standard CV benchmarks.} Table~\ref{tab:main_clean_ece} reports the calibration results on CIFAR-10, CIFAR-100 and ImageNet across five architectures. Overall, \textbf{our approach consistently achieves the lowest ECE,} outperforming all competing post-hoc baselines in all but one setting, where it ranks second. On CIFAR-10, \ours{} yields ECE of 0.29 (DN-121) and 0.31 (RN-50), yielding clear gains over the second-best method overall, i.e., FeatClip, across CNN and transformer backbones. On CIFAR-100, improvements are more pronounced, where \ours{} consistently outperforms the FeatClip by margins up to 0.45 (RN-50) and 0.38 (RN-18). Similarly, on ImageNet, \ours\;achieves the best calibration across all architectures, \textbf{reducing ECE by up to 0.44} (ViT-B) compared to FeatClip.

\noindent \textbf{B. Long-tailed distributions (Table~\ref{tab:cifar100_lt}).} We now report the calibration results under class-imbalance conditions, including CIFAR-100-LT with \textit{mild} (R-10) and \textit{extreme} (R-100) imbalance. 
In these scenarios, miscalibration is substantially amplified due to the skewed class-frequency distribution, where dense confidence regions are dominated by head-class predictions. Standard scalar calibration methods (e.g., TS), 
and recent strategies such as AdaTS and GC 
struggle to correct this heterogeneous behavior, 
leading to notably large miscalibration errors. Indeed, in some cases, these approaches further deteriorate the baseline performance 
(e.g., results in ImageNet-LT), rendering them ineffective in these regimes. In contrast, \ours{} 
yields substantial improvements, often achieving 
\textbf{reductions of 80-90\%} (see AdaTS and GC results in R-10 and R-100 settings in CIFAR-100-LT). Moreover, while most competing methods degrade sharply as the imbalance increases from R‑10 to R‑100, \ours{} remains comparatively stable, underscoring its robustness to severe class imbalance. Finally, 
FeatClip emerges as a competitive alternative in long-tailed scenarios, ranking second in most settings and first in four cases. However, we must highlight two important observations. First, when \ours{} ranks second, the ECE gap is marginal. And second, as shown in Fig.~\ref{fig:accuracy}, FeatClip’s accuracy is substantially degraded, sometimes \textbf{decreasing by 4\%-6\% compared to \ours,} highlighting \ours's ability to 
enhance calibration without compromising accuracy.



\begin{table}[!t]
\scriptsize
\centering
\caption{\textbf{Calibration performance in long-tailed image classification.}} 
\label{tab:cifar100_lt}
\begin{tabular}{llP{1cm}P{1cm}P{1cm}P{1cm}P{1cm}P{1.3cm}P{1.3cm}}
\toprule
\textbf{Dataset} & \textbf{Backbone} & \textbf{Uncal} & \textbf{TS} & \textbf{IR} &  \textbf{AdaTS} & \textbf{GC} &\textbf{FeatClip$^\ddagger$} & \textbf{\ours} \\
\midrule
\multirow{5}{*}{R-10 (mild)}
 & RN-50  & 19.53 & 19.23 & \underline{5.37} &  16.68 & 8.05 &6.62 & \textbf{3.76}\imp{-2.86} \\
 & RN-18  & 19.11 & 18.20 & 5.31 &  15.43 & 6.78 &\underline{4.73} & \textbf{3.54}\imp{-1.19} \\
 & DN-121 & 19.13 & 18.24 & 5.40 &  15.83 & 7.38 &\underline{6.59} & \textbf{3.60}\imp{-2.99} \\
 & ViT-S  & 19.14 & 19.07 & 5.80 &  16.75 & 8.36 &\textbf{3.19} & $\underline{3.40}_{+0.21}$ \\
 & ViT-B  & 24.12 & 24.18 & 5.24 &  21.91 & 12.60 &\underline{3.66} & \textbf{3.50}\imp{0.16} \\
\midrule

\multirow{5}{*}{R-100 (extreme)}
 & RN-50  & 48.64 & 44.27 & 27.20 &  34.74 & 25.84 &\textbf{4.78} & $\underline{5.06}_{+0.28}$ \\
 & RN-18  & 45.80 & 40.50 & 25.89 & 33.74 & 21.51 & \underline{2.37} & \textbf{2.30}\imp{-0.07} \\
 & DN-121 & 47.27 & 42.45 & 26.86 &  34.12 & 23.63 &\underline{4.87} & \textbf{3.07}\imp{-1.80} \\
 & ViT-S  & 50.24 & 46.41 & 30.71 & 35.55 & 29.74 & \underline{9.73} & \textbf{8.19}\imp{-1.54} \\
 & ViT-B  & 52.51 & 49.15 & 33.82 &  37.13 & 32.89 &\textbf{8.34} & $\underline{8.37}_{+0.03}$ \\
\midrule

\multirow{5}{*}{ImageNet-LT}
 & RN-50  & 3.71 & 4.16 & 4.50 &  3.72 & 4.03 &\underline{3.35} & \textbf{2.68}\imp{-0.67} \\
 & RN-18  & 2.61 & 2.86 & 2.56 &  2.56 & 4.06 &\underline{2.55} & \textbf{2.01}\imp{-0.54} \\
 & DN-121 & 2.61 & 2.73 & 2.75 &  2.42 & 2.65 &\textbf{1.51} & $\underline{1.62}_{+0.11}$ \\
 & ViT-S  & 2.81 & 3.51 & 2.71 &  2.76 & 2.97 &\underline{1.96} & \textbf{1.93}\imp{-0.03} \\
 & ViT-B  & 5.77 & 5.28 & 5.66 &  5.61 & 5.60 & \underline{5.59} & \textbf{3.59}\imp{-2.00} \\
\bottomrule
\end{tabular}%
\end{table}

\begin{table*}[h!]
\centering
\scriptsize
\caption{\textbf{Calibration performance under distributional drifts (CIFAR100 used for calibration, with ViTB-16 as backbone).} Note that HFE \cite{kim2024uncertainty}$^\star$ needs access to OOD samples of each target domain. 
Extended results in \App, Table~\ref{tab:cifar100c_ece_vitb16_full}.} 
\label{tab:cifar100_ds_summary}
\begin{tabular}{cP{1cm}P{1cm}P{1cm}P{1cm}P{1cm}P{1.3cm}P{1.3cm}|P{1.cm}}
\toprule
\textit{OOD-free?} & \ding{52} & \ding{52} & \ding{52} & \ding{52} & \ding{52} & \ding{52} & \ding{52} & \ding{56} \\ \midrule
 Severity &
\textbf{Uncal} & \textbf{TS} & \textbf{IR} & \textbf{AdaTS} & \textbf{GC} & \textbf{FeatClip}$^{\ddagger}$ & \textbf{\ours}  & \textbf{HFE}$^\star$ \\
\midrule

1  & 15.66	& 13.85	 & 13.78	& 13.61	& 10.08	& \underline{3.69} 
	& \bf 3.35\imp{-0.34} & 5.44\\
 2  & 19.00	&16.79	&14.40	&16.81	& 12.36 	&\underline{4.69}&	\bf 4.24\imp{-0.45} &7.34\\
 3 & 22.55	&19.92	&15.63	&19.98 & 14.77		&\underline{5.78}	& \bf 5.42\imp{-0.36} &9.42\\
 4  & 26.69	& 23.64	& 17.54	& 23.91& 17.88& \underline{7.70}	& \bf 7.20\imp{-0.50} &	12.12	 \\
 5  & 32.96	& 29.27	& 20.55	& 29.50	& 22.60 &	\underline{10.65}	& \bf 9.89\imp{-0.76} & 15.62 \\
\bottomrule
\end{tabular}%
\end{table*}
\begin{figure*}[!t]
    \centering
    \includegraphics[width=1.04\textwidth]{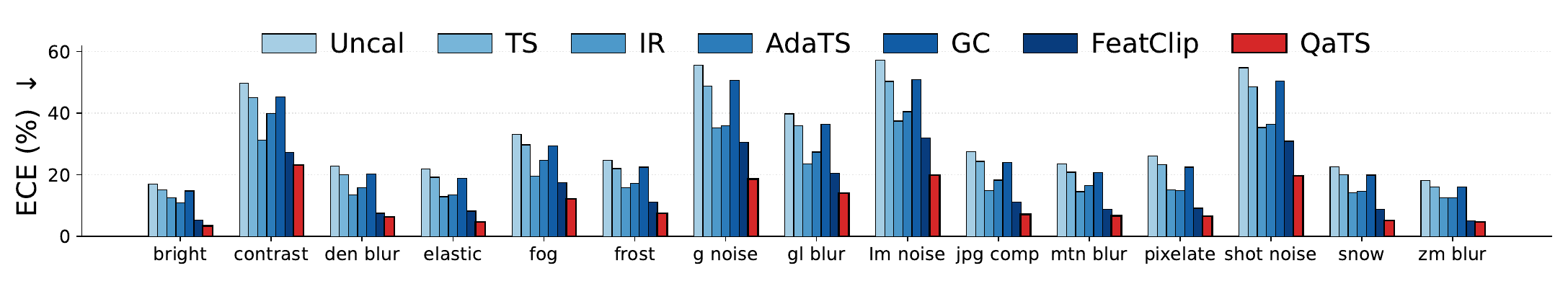}
    \vspace{-6mm}
    \caption{
    \textbf{Calibration under severe domain shift ( CIFAR-100-C, severity 5).} Per‑corruption ECE across all corruption types, highlighting the robustness of \ours{}. 
    }
    \label{fig:cifar100c_sev5}
\end{figure*}

\noindent \textbf{C. Performance under distributional drifts.} When considering real-world deployment, effective calibration methods should demonstrate robustness in handling samples from unknown distributions. To assess this capability in a controlled and reproducible manner, we now evaluate the performance of all methods on CIFAR-100C, a standard benchmark that introduces a wide range of synthetic corruptions designed to emulate realistic domain shifts such as noise, blur, weather effects, and digital artifacts. To do this, the QaTS parameters $(a,b)$, as well as the parameters for other approaches, are learned on the clean CIFAR-100 validation set and directly applied, without re-estimation, to the corrupted CIFAR-100C samples. We also include HFE \cite{kim2024uncertainty} as additional baseline, which is 
specifically designed to operate under distributional drifts. A key limitation of HFE, 
however, is its reliance on 
out‑of‑distribution (OOD) samples for every new target domain, an infeasible requirement in practice because such shifts are typically unknown in advance and cannot be exhaustively anticipated. Table~\ref{tab:cifar100_ds_summary} reports average ECE across corruptions for each severity level. Consistent with other settings, \ours{} achieves the \textbf{lowest ECE among all \textit{OOD-free} methods across all severity levels}. Notably, our approach substantially reduces ECE scores compared to all \textit{accuracy-preserving} methods, \textbf{yielding 2$\boldsymbol{\times}$}-\textbf{3$\boldsymbol{\times}$ lower ECE} values. In addition, Fig~\ref{fig:cifar100c_sev5} further depicts the ECE by corruption type for severity level 5, demonstrating the superiority of \ours{} across corruption types. 
Concerning \textit{non-accuracy preserving} methods, while performance gap compared to FeatClip is smaller, it widens as corruption severity increases, indicating stronger robustness to severe noise. 
Finally, even though HFE benefits from access to additional target-domain data, 
it still exhibits worse calibration 
than \ours{} (Table~\ref{tab:cifar100_ds_summary}, \textit{last column}), underscoring that our method provides a more effective and practical solution without 
requiring access to any target samples.

\noindent \textbf{D. Generalization to other tasks and domains.} We now assess the performance of \ours{} in a broader variety of domains and tasks, detailed below.

\noindent \textit{$\bullet$ \textbf{Medical Image classification.}} We first include experiments on medical image classification, a domain where calibrated confidence estimates are especially important for reliable clinical decision‑making. Average ECE results across 11 datasets for different CNN and ViT backbones are reported in Table~\ref{tab:medical_average}. Across all backbones, \textbf{\ours{} consistently achieves the lowest average ECE}, demonstrating robust calibration improvements also in this critical domain. 



\begin{table}[t!]
\centering
\scriptsize
\caption{\textbf{Calibration performance in medical image classification.} Average results across 10 datasets, whose extended results are reported in \App, Table \ref{tab:ece_miccai}.}
\begin{tabular}{lP{1.cm}P{1.cm}P{1.cm}P{1.cm}P{1.cm}P{1.2cm}P{1.2cm}}
\toprule
Backbone & Uncal & TS & IR & AdaTS & GC & FeatClip$^\ddagger$ & \bf \ours \\
\midrule
RN-50 & 7.53&	6.87&	6.42&	6.38&	\underline{4.98} &5.54&	\bf 3.56\imp{-1.42} \\
RN-18 & 10.64&	9.69&	7.53&	7.25&	\underline{5.96} & 6.67&	\bf4.03\imp{-1.93} \\
DN-121 & 10.09	&9.08	&7.14	&7.17& \underline{5.45}	&5.46	&\bf3.56\imp{-1.89} \\
ViT-S & 7.78	&7.22	&7.13	&7.12& 6.06	&\underline{4.42}	&\bf3.68\imp{-0.74} \\
ViT-B & 6.58	&5.91	&5.73	&5.97& 4.96&	\underline{3.91}	&\bf3.23\imp{-0.68} \\
\bottomrule
\end{tabular}
\label{tab:medical_average}
\end{table}

\noindent \textit{$\bullet$ \textbf{Image segmentation and Text classification (Table~\ref{tab:cross_domain}).}} Last, to highlight the broad applicability of \ours{}, we also assess its performance on a dense predictive task (i.e., semantic segmentation), and a non‑vision task (i.e., text classification). It can be observed  that the trend is similar to what we observed in image classification tasks, with \textbf{\ours{} outperforming all the baselines in both tasks}. Furthermore, similar to other scenarios, while FeatClip is competitive in terms calibration, it still degrades the discriminative performance of \textit{accuracy preserving} methods: 69.78$\rightarrow$69.51 
mIoU in segmentation, and 84.9$\rightarrow$84.7 accuracy in text classification. 
Thus, this comprehensive set of experiments across multiple scenarios, domains and tasks demonstrates that 
\ours{} remains effective across diverse scopes, emerging as a universal 
calibration strategy. 

\begin{table}[t!]
\centering
\scriptsize
\caption{\textbf{Calibration performance (ECE$\downarrow$) on additional tasks:} image semantic segmentation (Pascal VOC-2012) and text classification (NewsGroup-20). 
}
\label{tab:cross_domain}
\begin{tabular}{p{2.1cm}lP{0.9cm}P{0.9cm}P{0.9cm}P{0.9cm}P{0.9cm}P{1.1cm}P{1.1cm}}
\toprule
\textbf{Dataset} & \textbf{Backbone} & \textbf{Uncal} & \textbf{TS} & \textbf{IR} & \textbf{AdaTS} & \textbf{GC} & \textbf{FeatClip}$^\ddagger$ & \bf \ours \\
\midrule

VOC-2012 
& DLV3-R50 
& 4.49 & 3.69 & 3.23 & 3.49 & 3.26 & \underline{3.20} & \textbf{2.94}\imp{-0.26} \\
\midrule
NewsGroup-20 
& BERT-B 
& 5.81 & 5.42 & 4.73 & 4.80 & 4.53 & \underline{4.12} & \textbf{3.80}\imp{-0.32} \\

\bottomrule
\end{tabular}
\end{table}

\begin{figure}[!b]
    \centering
    \includegraphics[width=\linewidth]{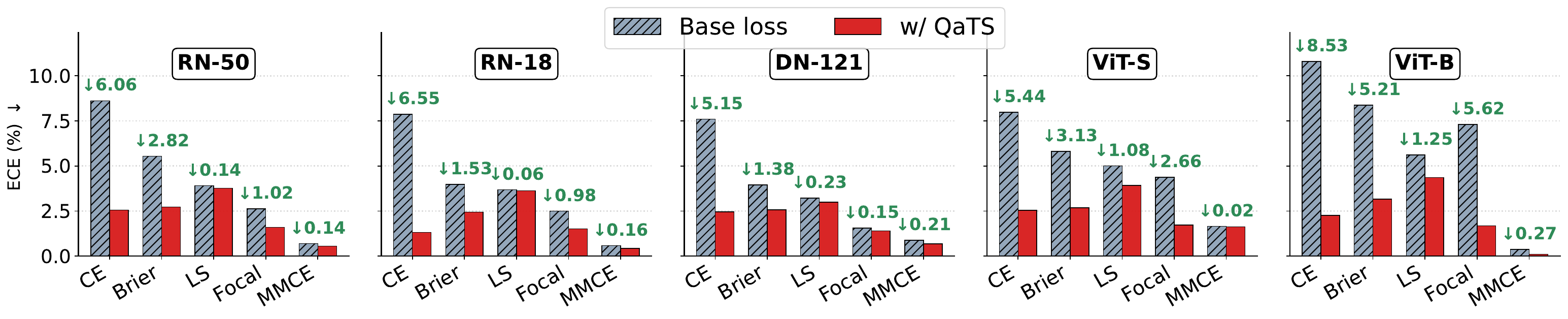}
    \caption{\textbf{QaTS complements training-time calibration.} ECE (\%, $\downarrow$) on CIFAR-100 (clean) across backbones and training losses (CE, Brier, LS-0.05, Focal, MMCE), comparing the base loss vs.\ adding QaTS post-hoc calibration.}
    \label{fig:diff_loss}
\end{figure}

\noindent \textbf{E. Compatibility with training-time calibration.} We also examine whether \ours{} compensates for standard trained models or also complements train-time regularization approaches. We evaluate \ours{} on models trained with calibration-aware objectives, including Brier loss \cite{glenn1950verification}, Label Smoothing (LS) \cite{szegedy2016rethinking} (with $\alpha=0.05$), Focal Loss \cite{lin2017focal}, and MMCE \cite{kumar2018trainable}, which are explicitly designed to reduce overconfidence 
during optimization. As shown in Fig.~\ref{fig:diff_loss}, while these losses generally reduce ECE compared to standard cross-entropy (CE), substantial miscalibration still remains. Applying \ours{} consistently yields further improvements across all backbones. For example, on CIFAR-100 with RN-50, Brier reduces ECE from 8.62 (CE) to 5.54, yet \ours{} further lowers it to 2.72. 
These results demonstrate that \ours{} acts as a complementary post-hoc approach that can be seamlessly applied on top of diverse training objectives without modification.

\begin{wraptable}{r}{0.46\textwidth}
\vspace{-20pt}
\centering
\scriptsize
\caption{Ablation on temperature function parameterizations (ViTB-16).}
\label{tab:functionals}
\setlength{\tabcolsep}{3.5pt}
\begin{tabular}{lccc}
\toprule
Method & C100 & C100-LT & C100-C \\
\midrule
Uncal & 10.79 & 52.51 & 32.96 \\
Linear (\ours{}) & 2.26 & \bf8.37 & \bf 9.89 \\
PW ($K$=2) & 2.25 & 8.44 & 10.14 \\
PW ($K$=4) & \bf 2.24 & 8.68 & 10.28 \\
PW ($K$=10) & 2.25 & 8.76 & 10.64 \\
\bottomrule
\end{tabular}
\vspace{-16pt}
\end{wraptable}
\noindent \textbf{F. Ablation on the temperature parameterization (Table~\ref{tab:functionals}).} While our default choice for the temperature function $T(q)$ is a simple linear form, more flexible monotone parameterizations 
could be used. 
To understand whether this additional flexibility yields measurable benefits, we replace the monotonic linear function in Eq. \ref{eq:temp_ours} with 
a piecewise (PW) linear temperature function defined over $K$ quantile segments (details in \App, \textsection \ref{sec:piecewise}). 
While the piecewise model can slightly reduce ECE on clean datasets, it yields worse results in long-tailed and distributional drifts scenarios. In long-tailed settings, where the calibration set is highly skewed, the larger number of parameters (i.e., $K+1$) tends to overfit the sparsely populated quantile regions, leading to slightly worse generalization. Under distributional drifts, the same higher capacity makes the model more sensitive to confidence patterns that differ from those seen 
during calibration, 
reducing its robustness. This behavior is reflected empirically by the gradual increase of the ECE as $K$ grows. In contrast, the linear form adopted in \ours{}, with only two parameters, consistently provides the best robustness–complexity trade‑off, capturing the smooth trend of miscalibration across quantiles while avoiding overfitting under complex scenarios. We stress that these observations concern only the choice of the parameterization for $T(q)$ and the underlying quantile‑based calibration principle of \ours{} remains the same across all monotone function classes.





\section{Conclusion}
\label{sec:conclusion}

In this work, we show that miscalibration is highly heterogeneous across the confidence spectrum and shifts markedly across standard and more complex regimes, such as long-tailed. This variability exposes a core limitation of existing smooth temperature mappings, which are based on raw-confidence scores and lack an explicit mechanism to target the quantile regions where errors are concentrated. \ours{} addresses this by operating directly in quantile space, allowing the temperature to depend on the \textit{relative} position of a prediction within the calibration distribution. This rank‑based formulation naturally aligns with a reparameterized ECE objective and remains stable across a wide range of conditions. Evaluated over diverse datasets, architectures, and tasks, this principle yields consistently improved calibration while preserving accuracy, offering a simple and broadly applicable foundation for reliable post‑hoc calibration.

\bibliographystyle{splncs04}
\bibliography{ECCV_26/main_ref}

@String(CVPR  = {IEEE Conf. Comput. Vis. Pattern Recog.})

@String(ICCV  = {Int. Conf. Comput. Vis.})

@String(NeurIPS = {Adv. Neural Inform. Process. Syst.})

@String(ICML  = {Int. Conf. Mach. Learn.})

@String(ICLR  = {Int. Conf. Learn. Represent.})

@String(AAAI  = {AAAI})

@String(CVPR  = {CVPR})

@String(ICCV  = {ICCV})

@String(NeurIPS = {NeurIPS})

@String(ICML  = {ICML})

@String(ICLR  = {ICLR})

@inproceedings{tomani2022parameterized,
  title={Parameterized temperature scaling for boosting the expressive power in post-hoc uncertainty calibration},
  author={Tomani, Christian and Cremers, Daniel and Buettner, Florian},
  booktitle={European conference on computer vision},
  pages={555--569},
  year={2022}
}

@mastersthesis{krizhevsky2009cifar,
	title        = {{Learning Multiple Layers of Features from Tiny Images}},
	author       = {Alex Krizhevsky and Vinod Nair and Geoffrey Hinton},
	year         = 2009,
	url          = {https://www.cs.toronto.edu/~kriz/learning-features-2009-TR.pdf},
	school       = {University of Toronto}
}

@article{russakovsky2015imagenet,
  title={{Imagenet Large Scale Visual Recognition Challenge}},
  author={Russakovsky, Olga and Deng, Jia and Su, Hao and Krause, Jonathan and Satheesh, Sanjeev and Ma, Sean and Huang, Zhiheng and Karpathy, Andrej and Khosla, Aditya and Bernstein, Michael and others},
  journal={International journal of computer vision},
  volume={115},
  number={3},
  pages={211--252},
  year={2015},
  publisher={Springer}
}

@misc{twenty_newsgroups_113,
  title={Twenty Newsgroups [Dataset]. UCI Machine Learning Repository},
  author={Mitchell, T},
  year={1997}
}

@inproceedings{cao2019learning,
	title        = {{Learning imbalanced datasets with label-distribution-aware margin loss}},
	author       = {Cao, Kaidi and Wei, Colin and Gaidon, Adrien and Arechiga, Nikos and Ma, Tengyu},
	year         = 2019,
	booktitle    = NeurIPS,
	url          = {https://arxiv.org/abs/1906.07413}
}

@inproceedings{liu2019imagenetlt,
	title        = {{Large-Scale Long-Tailed Recognition in an Open World}},
	author       = {Ziwei Liu and Zhongqi Miao and Xiaohang Zhan and Jiayun Wang and Boqing Gong and Stella X. Yu},
	year         = 2019,
	booktitle    = CVPR
}

@inproceedings{joy2023sample,
  title={Sample-dependent adaptive temperature scaling for improved calibration},
  author={Joy, Tom and Pinto, Francesco and Lim, Ser-Nam and Torr, Philip HS and Dokania, Puneet K},
  booktitle={Proceedings of the AAAI Conference on Artificial Intelligence},
  volume={37},
  pages={14919--14926},
  year={2023}
}

@inproceedings{hekler2023test,
  title={Test time augmentation meets post-hoc calibration: uncertainty quantification under real-world conditions},
  author={Hekler, Achim and Brinker, Titus J and Buettner, Florian},
  booktitle={Proceedings of the AAAI Conference on Artificial Intelligence},
  pages={14856--14864},
  year={2023}
}

@inproceedings{liu2022devil,
  title={The devil is in the margin: Margin-based label smoothing for network calibration},
  author={Liu, Bingyuan and Ben Ayed, Ismail and Galdran, Adrian and Dolz, Jose},
  booktitle={Proceedings of the IEEE/CVF Conference on Computer Vision and Pattern Recognition (CVPR)},
  pages={80--88},
  year={2022}
}

@inproceedings{guo2017calibration,
  title={On calibration of modern neural networks},
  author={Guo, Chuan and Pleiss, Geoff and Sun, Yu and Weinberger, Kilian Q},
  booktitle={International conference on machine learning (ICML)},
  pages={1321--1330},
  year={2017},
  organization={PMLR}
}

@inproceedings{ding2021local,
  title={Local temperature scaling for probability calibration},
  author={Ding, Zhipeng and Han, Xu and Liu, Peirong and Niethammer, Marc},
  booktitle={Proceedings of the IEEE/CVF International Conference on Computer Vision},
  pages={6889--6899},
  year={2021}
}

@article{ovadia2019can,
  title={Can you trust your model's uncertainty? evaluating predictive uncertainty under dataset shift},
  author={Ovadia, Yaniv and Fertig, Emily and Ren, Jie and Nado, Zachary and Sculley, David and Nowozin, Sebastian and Dillon, Joshua and Lakshminarayanan, Balaji and Snoek, Jasper},
  journal={Advances in neural information processing systems (NeurIPS)},
  volume={32},
  year={2019}
}

@inproceedings{pereyra2017regularizing,
  title={Regularizing Neural Networks by Penalizing Confident Output Distributions},
  author={Pereyra, Gabriel and Tucker, George and Chorowski, Jan and Kaiser, Lukasz and Hinton, Geoffrey},
  booktitle={International Conference on Learning Representations (ICLR)},
  year={2017}
}

@article{mukhoti2020calibrating,
  title={Calibrating deep neural networks using focal loss},
  author={Mukhoti, Jishnu and Kulharia, Viveka and Sanyal, Amartya and Golodetz, Stuart and Torr, Philip and Dokania, Puneet},
  journal={Advances in Neural Information Processing Systems (NeurIPS)},
  volume={33},
  pages={15288--15299},
  year={2020}
}

@inproceedings{park2023acls,
  title={Acls: Adaptive and conditional label smoothing for network calibration},
  author={Park, Hyekang and Noh, Jongyoun and Oh, Youngmin and Baek, Donghyeon and Ham, Bumsub},
  booktitle={Proceedings of the IEEE/CVF International Conference on Computer Vision (ICCV)},
  pages={3936--3945},
  year={2023}
}

@inproceedings{cheng2022calibrating,
  title={Calibrating deep neural networks by pairwise constraints},
  author={Cheng, Jiacheng and Vasconcelos, Nuno},
  booktitle={Proceedings of the IEEE/CVF Conference on Computer Vision and Pattern Recognition (CVPR)},
  pages={13709--13718},
  year={2022}
}

@inproceedings{tao2025feature,
  title={Feature clipping for uncertainty calibration},
  author={Tao, Linwei and Dong, Minjing and Xu, Chang},
  booktitle={Proceedings of the AAAI Conference on Artificial Intelligence},
  pages={20841--20849},
  year={2025}
}

@inproceedings{guptacalibration,
  title={Calibration of Neural Networks using Splines},
  author={Gupta, Kartik and Rahimi, Amir and Ajanthan, Thalaiyasingam and Mensink, Thomas and Sminchisescu, Cristian and Hartley, Richard},
  booktitle={International Conference on Learning Representations},
year={2021}
}

@inproceedings{zhang2020mix,
  title={Mix-n-match: Ensemble and compositional methods for uncertainty calibration in deep learning},
  author={Zhang, Jize and Kailkhura, Bhavya and Han, T Yong-Jin},
  booktitle={International conference on machine learning},
  pages={11117--11128},
  year={2020},
  organization={PMLR}
}

@article{medmnistv2,
    title={Med{MNIST} v2-{A} large-scale lightweight benchmark for 2{D} and 3{D} biomedical image classification},
    author={Yang, Jiancheng and Shi, Rui and Wei, Donglai and Liu, Zequan and Zhao, Lin and Ke, Bilian and Pfister, Hanspeter and Ni, Bingbing},
    journal={Scientific Data},
    volume={10},
    number={1},
    pages={41},
    year={2023},
    publisher={Nature Publishing Group UK London}
}

@inproceedings{gong2021confidence,
  title={Confidence calibration for domain generalization under covariate shift},
  author={Gong, Yunye and Lin, Xiao and Yao, Yi and Dietterich, Thomas G and Divakaran, Ajay and Gervasio, Melinda},
  booktitle={Proceedings of the IEEE/CVF international conference on computer vision},
  pages={8958--8967},
  year={2021}
}

@article{yang2023beyond,
  title={Beyond probability partitions: Calibrating neural networks with semantic aware grouping},
  author={Yang, Jia-Qi and Zhan, De-Chuan and Gan, Le},
  journal={Advances in Neural Information Processing Systems},
  volume={36},
  pages={58448--58460},
  year={2023}
}

@inproceedings{medmnistv1,
    title={Med{MNIST} Classification Decathlon: A Lightweight Auto{ML} Benchmark for Medical Image Analysis},
    author={Yang, Jiancheng and Shi, Rui and Ni, Bingbing},
    booktitle={IEEE 18th International Symposium on Biomedical Imaging (ISBI)},
    pages={191--195},
    year={2021}
}

@inproceedings{jung2023scaling,
  title={Scaling of class-wise training losses for post-hoc calibration},
  author={Jung, Seungjin and Seo, Seungmo and Jeong, Yonghyun and Choi, Jongwon},
  booktitle={International Conference on Machine Learning},
  pages={15421--15434},
  year={2023},
  organization={PMLR}
}

@article{platt1999probabilistic,
  title={Probabilistic outputs for support vector machines and comparisons to regularized likelihood methods},
  author={Platt, John and others},
  journal={Advances in large margin classifiers},
  volume={10},
  number={3},
  pages={61--74},
  year={1999},
  publisher={Cambridge, MA}
}

@article{minderer2021revisiting,
  title={Revisiting the calibration of modern neural networks},
  author={Minderer, Matthias and Djolonga, Josip and Romijnders, Rob and Hubis, Frances and Zhai, Xiaohua and Houlsby, Neil and Tran, Dustin and Lucic, Mario},
  journal={Advances in neural information processing systems},
  volume={34},
  pages={15682--15694},
  year={2021}
}

@inproceedings{larrazabal2023maximum,
  title={Maximum entropy on erroneous predictions: Improving model calibration for medical image segmentation},
  author={Larrazabal, Agostina J and Mart{\'\i}nez, C{\'e}sar and Dolz, Jose and Ferrante, Enzo},
  booktitle={International conference on medical image computing and computer-assisted intervention},
  pages={273--283},
  year={2023},
  organization={Springer}
}

@inproceedings{kim2024uncertainty,
  title={Uncertainty calibration with energy based instance-wise scaling in the wild dataset},
  author={Kim, Mijoo and Kwon, Junseok},
  booktitle={European Conference on Computer Vision},
  pages={232--248},
  year={2024}
}

@inproceedings{tomani2021post,
  title={Post-hoc uncertainty calibration for domain drift scenarios},
  author={Tomani, Christian and Gruber, Sebastian and Erdem, Muhammed Ebrar and Cremers, Daniel and Buettner, Florian},
  booktitle={Proceedings of the IEEE/CVF Conference on Computer Vision and Pattern Recognition},
  pages={10124--10132},
  year={2021}
}

@article{yu2022robust,
  title={Robust calibration with multi-domain temperature scaling},
  author={Yu, Yaodong and Bates, Stephen and Ma, Yi and Jordan, Michael},
  journal={Advances in Neural Information Processing Systems},
  volume={35},
  pages={27510--27523},
  year={2022}
}

@inproceedings{liu2023class,
  title={Class Adaptive Network Calibration},
  author={Liu, Bingyuan and Rony, J{\'e}r{\^o}me and Galdran, Adrian and Dolz, Jose and Ben Ayed, Ismail},
  booktitle={Proceedings of the IEEE/CVF Conference on Computer Vision and Pattern Recognition (CVPR)},
  pages={16070--16079},
  year={2023}
}

@inproceedings{murugesan2023trust,
  title={Trust your neighbours: Penalty-based constraints for model calibration},
  author={Murugesan, Balamurali and Adiga Vasudeva, Sukesh and Liu, Bingyuan and Lombaert, Herv{\'e} and Ben Ayed, Ismail and Dolz, Jose},
  booktitle={International Conference on Medical Image Computing and Computer-Assisted Intervention (MICCAI)},
  pages={572--581},
  year={2023}
}

@article{hendrycks2019robustness,
  title={Benchmarking Neural Network Robustness to Common Corruptions and Perturbations},
  author={Dan Hendrycks and Thomas Dietterich},
  journal={Proceedings of the International Conference on Learning Representations},
  year={2019}
}

@article{everingham2015pascal,
  title={The pascal visual object classes challenge: A retrospective},
  author={Everingham, Mark and Eslami, SM Ali and Van Gool, Luc and Williams, Christopher KI and Winn, John and Zisserman, Andrew},
  journal={International journal of computer vision},
  volume={111},
  number={1},
  pages={98--136},
  year={2015},
  publisher={Springer}
}

@article{tschandl2018ham10000,
  title={The HAM10000 dataset, a large collection of multi-source dermatoscopic images of common pigmented skin lesions},
  author={Tschandl, Philipp and Rosendahl, Cliff and Kittler, Harald},
  journal={Scientific data},
  volume={5},
  number={1},
  pages={180161},
  year={2018},
  publisher={Nature Publishing Group}
}

@inproceedings{zadrozny2002transforming,
  title={Transforming classifier scores into accurate multiclass probability estimates},
  author={Zadrozny, Bianca and Elkan, Charles},
  booktitle={Proceedings of the eighth ACM SIGKDD international conference on Knowledge discovery and data mining},
  pages={694--699},
  year={2002}
}

@article{chen2017rethinking,
  title={Rethinking atrous convolution for semantic image segmentation},
  author={Chen, Liang-Chieh and Papandreou, George and Schroff, Florian and Adam, Hartwig},
  journal={arXiv preprint arXiv:1706.05587},
  year={2017}
}

@inproceedings{devlin2019bert,
  title={Bert: Pre-training of deep bidirectional transformers for language understanding},
  author={Devlin, Jacob and Chang, Ming-Wei and Lee, Kenton and Toutanova, Kristina},
  booktitle={Proceedings of the 2019 conference of the North American chapter of the association for computational linguistics: human language technologies, volume 1 (long and short papers)},
  pages={4171--4186},
  year={2019}
}

@article{muller2019does,
  title={When does label smoothing help?},
  author={M{\"u}ller, Rafael and Kornblith, Simon and Hinton, Geoffrey E},
  journal={Advances in neural information processing systems (NeurIPS)},
  volume={32},
  year={2019}
}

@inproceedings{zhang2022and,
  title={When and how mixup improves calibration},
  author={Zhang, Linjun and Deng, Zhun and Kawaguchi, Kenji and Zou, James},
  booktitle={International Conference on Machine Learning},
  pages={26135--26160},
  year={2022},
  organization={PMLR}
}

@article{thulasidasan2019mixup,
  title={On mixup training: Improved calibration and predictive uncertainty for deep neural networks},
  author={Thulasidasan, Sunil and Chennupati, Gopinath and Bilmes, Jeff A and Bhattacharya, Tanmoy and Michalak, Sarah},
  journal={Advances in Neural Information Processing Systems (NeurIPS)},
  volume={32},
  year={2019}
}

@inproceedings{niculescu2005obtaining,
  title={Obtaining calibrated probability estimates from decision trees and naive Bayesian classifiers},
  author={Zadrozny, Bianca and Elkan, Charles},
  booktitle={International Conference on Machine Learning},
  pages={625--632},
  year={2001}
}

@article{xiong2023proximity,
  title={Proximity-informed calibration for deep neural networks},
  author={Xiong, Miao and Deng, Ailin and Koh, Pang Wei W and Wu, Jiaying and Li, Shen and Xu, Jianqing and Hooi, Bryan},
  journal={Advances in Neural Information Processing Systems},
  volume={36},
  pages={68511--68538},
  year={2023}
}

@inproceedings{frenkel2021network,
  title={Network calibration by class-based temperature scaling},
  author={Frenkel, Lior and Goldberger, Jacob},
  booktitle={2021 29th European Signal Processing Conference (EUSIPCO)},
  pages={1486--1490},
  year={2021},
  organization={IEEE}
}

@inproceedings{murugesan2024robust,
  title={Robust calibration of large vision-language adapters},
  author={Murugesan, Balamurali and Silva-Rodr{\'\i}guez, Julio and Ayed, Ismail Ben and Dolz, Jose},
  booktitle={European Conference on Computer Vision},
  pages={147--165},
  year={2024},
  organization={Springer}
}

@inproceedings{wei2022mitigating,
  title={Mitigating neural network overconfidence with logit normalization},
  author={Wei, Hongxin and Xie, Renchunzi and Cheng, Hao and Feng, Lei and An, Bo and Li, Yixuan},
  booktitle={International Conference on Machine Learning (ICML)},
  pages={23631--23644},
  year={2022},
  organization={PMLR}
}

@inproceedings{szegedy2016rethinking,
  title={{Rethinking the inception architecture for computer vision}},
  author={Szegedy, Christian and Vanhoucke, Vincent and Ioffe, Sergey and Shlens, Jon and Wojna, Zbigniew},
  booktitle={Proceedings of the IEEE conference on computer vision and pattern recognition (CVPR)},
  pages={2818--2826},
  year={2016}
}

@inproceedings{lin2017focal,
  title={{Focal Loss for Dense Object Detection}},
  author={Lin, Tsung-Yi and Goyal, Priya and Girshick, Ross and He, Kaiming and Doll{\'a}r, Piotr},
  booktitle={Proceedings of the IEEE international conference on computer vision (ICCV)},
  pages={2980--2988},
  year={2017}
}

@inproceedings{larrazabal2021maximum,
  title={{Maximum Entropy on Erroneous Predictions ({MEEP}): Improving model calibration for medical image segmentation}},
  author={Larrazabal, Agostina and Martinez, Cesar and Dolz, Jose and Ferrante, Enzo},
  booktitle={International Conference on Medical Image Computing and Computer-Assisted Intervention (MICCAI)},
  year={2023}
}

@article{glenn1950verification,
  title={{Verification of Forecasts Expressed in Terms of Probability}},
  author={Glenn, W Brier and others},
  journal={Monthly weather review},
  volume={78},
  number={1},
  pages={1--3},
  year={1950},
  publisher={War Department, Office of the Chief Signal Officer}
}

@inproceedings{kumar2018trainable,
  title={{Trainable Calibration Measures for Neural Networks from Kernel Mean Embeddings}},
  author={Kumar, Aviral and Sarawagi, Sunita and Jain, Ujjwal},
  booktitle={International Conference on Machine Learning},
  pages={2805--2814},
  year={2018},
  organization={PMLR}
}


\title{Quantile‑Adaptive Temperature Scaling for Confidence Calibration \\ Supplementary Material}
\titlerunning{Supplementary Material}

\author{Anonymous ECCV submission}
\institute{}

\maketitle

\appendix



\section{Dataset Details}

We evaluate \ours{} across a diverse set of benchmarks covering standard image classification, long-tailed versions, distribution shifts and medical imaging datasets. Below we briefly summarize the datasets used in our experiments.

\paragraph{Standard image classification datasets.}
We first evaluate on standard image benchmarks. 
\textbf{CIFAR-10} and \textbf{CIFAR-100} consist of color images with 10 and 100 classes, respectively, and contain 50k training and 10k test samples~\cite{krizhevsky2009cifar}. 
We also evaluate on \textbf{ImageNet-1K}~\cite{russakovsky2015imagenet}, a large-scale dataset containing approximately 1.28M training images and 50k validation images across 1000 object categories. These datasets serve as standard benchmarks for evaluating calibration methods under balanced training distributions.

\paragraph{Long-tailed recognition datasets.}
To evaluate calibration under severe class imbalance, we conduct experiments on long-tailed recognition benchmarks derived from CIFAR-100 and ImageNet. 
Specifically, \textbf{CIFAR-100-LT} is constructed by subsampling the CIFAR-100 training set with an exponential decay across classes to produce imbalance ratios of $R=10$ and $R=100$~\cite{cao2019learning}. 
We also evaluate on \textbf{ImageNet-LT}~\cite{liu2019imagenetlt}, a long-tailed subset of ImageNet with naturally imbalanced class frequencies. 
These benchmarks simulate realistic scenarios where head classes contain many training examples while tail classes have only limited supervision.
\paragraph{Distribution shift benchmark.}
We also evaluate robustness to distributional shifts using \textbf{CIFAR-100-C}~\cite{hendrycks2019robustness}, which introduces synthetic corruptions to the CIFAR-100 test images. 
The benchmark contains 15 corruption types: \textit{gaussian noise}, \textit{shot noise}, \textit{impulse noise}, \textit{defocus blur}, \textit{glass blur}, \textit{motion blur}, \textit{zoom blur}, \textit{snow}, \textit{frost}, \textit{fog}, \textit{brightness}, \textit{contrast}, \textit{elastic transform}, \textit{pixelate}, and \textit{jpeg compression}, each applied at \emph{five} severity levels.

Following standard practice, calibration parameters are learned using the clean CIFAR-100 validation set and are then applied directly to corrupted samples without re-estimation.

\paragraph{Medical imaging datasets.}
To evaluate calibration performance in real-world scenarios like medical applications, we conduct experiments on 11 medical imaging datasets. 
These include \textbf{HAM10000}~\cite{tschandl2018ham10000}, a dermatoscopic skin lesion dataset, and ten datasets from the \textbf{MedMNIST} benchmark~\cite{medmnistv1,medmnistv2}: 
\emph{PathMNIST}, \emph{BloodMNIST}, \emph{BreastMNIST}, \emph{DermaMNIST}, \emph{OrganAMNIST}, \emph{OrganCMNIST}, \emph{OrganSMNIST}, \emph{PneumoniaMNIST}, \emph{RetinaMNIST}, and \emph{TissueMNIST}. 
These datasets cover a diverse set of medical modalities including histopathology, dermatology, hematology, organ CT imaging, retinal imaging, and chest X-ray classification. 
The MedMNIST datasets provide standardized splits designed for lightweight benchmarking of medical imaging methods.

\paragraph{Additional tasks.}
To further assess the generality of \ours{} beyond standard image classification, we also evaluate it on a text classification benchmark and a dense prediction benchmark.
\textbf{20 Newsgroups}~\cite{twenty_newsgroups_113} is a widely used natural language processing dataset for topic classification, containing documents from 20 different newsgroups.
For dense prediction, we use \textbf{Pascal VOC 2012}~\cite{everingham2015pascal}, a standard semantic segmentation benchmark with 21 semantic classes (including background).
These experiments allow us to evaluate whether the proposed calibration strategy generalizes across modalities and tasks beyond image classification.

Overall, the evaluation spans a total of \textbf{162} experimental settings across multiple datasets, model architectures (ResNet-18, ResNet-50, DenseNet-121, ViT-S/16, and ViT-B/16), distribution shifts and tasks, providing a comprehensive assessment of calibration performance in diverse real-world scenarios.


\section{Extended Results for Domain shift}

Table~\ref{tab:cifar100c_ece_vitb16_full} extends Table~3 of the main paper to provide the full CIFAR-100-C results for individual corruptions and severity. This detailed breakdown confirms the trends observed in the averaged results. First, calibration error increases steadily with corruption severity for all methods, highlighting the difficulty of maintaining reliable confidence estimates under progressively stronger distribution shifts. Second, while standard post-hoc baselines such as TS, IR and AdaTS improve over the uncalibrated model in many cases, their performance degrades substantially under severe corruptions, especially for noise-based perturbations such as \textit{gaussian noise}, \textit{impulse noise} and \textit{shot noise}.
In contrast, \ours{} consistently achieves the lowest overall ECE across all five severity levels among OOD-free methods, matching the conclusions drawn from Table~3 in the main paper. The corruption-wise results further show that these gains are not driven by only a few favorable cases: \ours{} remains strongest across all corruption types, particularly on challenging perturbations such as \textit{elastic transform}, \textit{gaussian blur}, \textit{gaussian noise}, \textit{jpeg compression} and \textit{zoom blur}.

A further important observation is that HFE benefits from access to additional OOD samples, whereas \ours{} does not. Despite this stronger supervision setting, \ours{} still yields lower average ECE at every severity level. Compared to FeatClip, the gap is smaller on certain corruptions, particularly at high severity, but \ours{} still achieves the best overall averages while preserving the original classifier accuracy, unlike FeatClip. 

\begin{table*}[!h]
\centering
\setlength{\tabcolsep}{3.2pt}
\renewcommand{\arraystretch}{1.08}
\begin{adjustbox}{width=\textwidth}
\begin{tabular}{c l c
                c c c c c c c c c c c c c c c
                c}
\toprule
\multirow{2}{*}{severity} & \multirow{2}{*}{method} & \multirow{2}{*}{$D_{\text{OOD}}$} &
\multicolumn{15}{c}{CIFAR-100-C (ECE \%) per corruption} & \multirow{2}{*}{\textbf{Average}} \\
\cmidrule(lr){4-18}
& & &
brightness & contrast & df blur & el transf & Fog & Frost & G noise & Gl blur & i noise & jpg comp & mtn blur & pixelate & shot\_noise & snow & zm blur &
 \\
\midrule

\multirow{8}{*}{1}
& Uncal    & \ding{56} & 11.17 & 11.01 & 10.90 & 13.16 & 11.03 & 13.03 & 25.36 & 30.98 & 20.24 & 17.91 & 12.43 & 11.94 & 18.22 & 14.37 & 13.23 & 15.66 \\
& TS       & \ding{56} &  9.93 &  9.78 &  9.68 & 11.67 &  9.80 & 11.53 & 22.50 & 27.76 & 17.32 & 15.83 & 10.97 & 10.61 & 16.08 & 12.70 & 11.65 & 13.85 \\
& IR       & \ding{56} & 11.38 & 13.78 & 14.16 & 12.82 & 13.93 & 13.08 & 15.48 & 18.61 & 12.03 & 12.69 & 12.92 & 13.68 & 13.39 & 13.50 & 12.28 & 13.58 \\
& AdaTS    & \ding{56} &  9.65 &  9.53 &  9.45 & 11.51 &  9.61 & 11.12 & 21.40 & 28.56 & 16.48 & 15.43 & 11.12 & 10.41 & 16.28 & 12.03 & 11.62 & 13.61 \\
& GC       & \ding{56} &  7.14 &  7.30 &  6.98 &  8.50 &  7.19 &  8.32 & 16.48 & 20.27 & 12.52 & 11.55 &  7.99 &  7.64 & 11.93 &  8.92 &  8.50 & 10.08 \\
& HFE      & \ding{52} &  3.28 &  3.97 &  3.37 &  3.89 &  4.00 &  3.59 & 10.51 & 14.99 &  5.38 &  6.04 &  3.85 &  3.87 &  6.35 &  4.91 &  3.66 &  5.44 \\
& FeatClip & \ding{56} &  \underline{2.82} &  \underline{2.86} &  \underline{2.70} &  \underline{3.18} &  \underline{2.80} &  \underline{2.88} &  \underline{7.08} &  \underline{9.01} &  \underline{2.27} &  \underline{3.84} &  \underline{2.97} &  \underline{2.85} &  \underline{4.24} &  \underline{2.94} &  \underline{2.94} &  \underline{3.69} \\
\rowcolor{lightgray}
& \ours    & \ding{56} & \textbf{2.33} & \textbf{2.24} & \textbf{2.42} & \textbf{2.63} & \textbf{2.53} & \textbf{2.44} & \textbf{6.09} & \textbf{9.54} & \textbf{1.82} & \textbf{3.37} & \textbf{2.63} & \textbf{2.86} & \textbf{3.79} & \textbf{3.05} & \textbf{2.45} & \textbf{3.35} \\
\midrule

\multirow{8}{*}{2}
& Uncal    & \ding{56} & 11.50 & 13.85 & 11.52 & 12.84 & 12.85 & 15.47 & 39.25 & 31.01 & 28.50 & 21.82 & 15.60 & 12.77 & 26.25 & 18.18 & 13.64 & 19.00 \\
& TS       & \ding{56} & 10.22 & 12.28 & 10.22 & 11.37 & 11.40 & 13.59 & 34.88 & 27.76 & 24.38 & 19.27 & 13.79 & 11.37 & 23.27 & 16.08 & 12.03 & 16.79 \\
& IR       & \ding{56} & 11.44 & 11.89 & 13.23 & 12.81 & 12.69 & 12.39 & 23.68 & 18.17 & 15.45 & 13.40 & 12.91 & 13.29 & 16.74 & 12.94 & 12.04 & 14.20 \\
& AdaTS    & \ding{56} & 10.13 & 12.01 & 10.07 & 11.25 & 10.96 & 13.59 & 36.26 & 28.30 & 24.07 & 18.84 & 13.67 & 11.24 & 24.22 & 15.71 & 11.85 & 16.81 \\
& GC       & \ding{56} &  7.29 &  9.71 &  7.41 &  8.26 &  8.58 &  9.83 & 25.47 & 20.33 & 17.95 & 14.20 & 10.37 &  8.22 & 17.30 & 11.45 &  8.99 & 12.36 \\
& HFE      & \ding{52} &  3.31 &  4.40 &  3.69 &  3.94 &  4.26 &  5.26 & 20.01 & 14.62 & 11.12 &  8.27 &  4.92 &  3.96 & 11.52 &  6.99 &  3.78 &  7.34 \\
& FeatClip & \ding{56} &  \underline{2.68} &  \underline{3.57} &  \underline{2.86} &  \underline{3.32} &  \underline{2.84} &  \underline{3.21} & \underline{12.16} &  \underline{9.26} &  \underline{2.78} &  \underline{5.27} &  \underline{4.25} &  \underline{3.33} &  \underline{7.27} &  \underline{4.21} &  \underline{3.30} &  \underline{4.69} \\
\rowcolor{lightgray}
& \ours    & \ding{56} & \textbf{2.47} & \textbf{2.86} & \textbf{2.27} & \textbf{2.53} & \textbf{2.37} & \textbf{3.08} & \textbf{11.89} & \textbf{9.65} & 3.44 & \textbf{4.35} & \textbf{3.38} & \textbf{2.50} & \textbf{6.31} & \textbf{4.07} & \textbf{2.40} & \textbf{4.24} \\
\midrule

\multirow{8}{*}{3}
& Uncal    & \ding{56} & 12.46 & 16.90 & 12.79 & 14.22 & 15.58 & 19.88 & 48.77 & 27.24 & 37.47 & 23.43 & 19.48 & 13.80 & 42.62 & 18.82 & 14.79 & 22.55 \\
& TS       & \ding{56} & 11.09 & 14.95 & 11.27 & 12.58 & 13.79 & 17.64 & 43.00 & 24.21 & 32.40 & 20.70 & 17.27 & 12.27 & 37.98 & 16.58 & 13.01 & 19.92 \\
& IR       & \ding{56} & 13.64 & 11.75 & 12.76 & 12.67 & 12.52 & 13.65 & 30.76 & 15.44 & 20.32 & 13.98 & 12.85 & 13.28 & 25.65 & 13.18 & 12.03 & 15.63 \\
& AdaTS    & \ding{56} & 10.80 & 14.93 & 11.17 & 12.39 & 13.32 & 17.46 & 44.83 & 24.16 & 32.19 & 20.36 & 17.15 & 11.90 & 39.85 & 16.41 & 12.86 & 19.98 \\
& GC       & \ding{56} &  7.95 & 12.12 &  8.18 &  9.18 & 10.58 & 13.23 & 31.13 & 17.47 & 24.71 & 15.33 & 13.38 &  8.88 & 28.22 & 11.42 &  9.78 & 14.77 \\
& HFE      & \ding{52} &  3.65 &  5.77 &  3.97 &  4.08 &  5.26 &  7.96 & 26.34 & 11.65 & 17.44 &  9.24 &  7.04 &  4.31 & 23.29 &  7.22 &  4.08 &  9.42 \\
& FeatClip & \ding{56} & \underline{2.78} & \underline{4.23} & \underline{3.00} & \underline{3.50} & \underline{3.30} & \underline{4.98} & \underline{16.05} & \underline{7.21} & \underline{6.58} & \underline{5.79} & \underline{4.92} & \underline{3.21} & \underline{14.13} & \underline{3.46} & \underline{3.58} & \underline{5.78} \\
\rowcolor{lightgray}
& \ours    & \ding{56} & \textbf{2.46} & \textbf{3.02} & \textbf{2.62} & \textbf{2.94} & \textbf{3.01} & \textbf{4.56} & \textbf{15.68} & \textbf{7.13} & 7.56 & \textbf{5.10} & \textbf{3.80} & \textbf{2.91} & \textbf{13.99} & \textbf{3.78} & \textbf{2.79} & \textbf{5.42} \\
\midrule

\multirow{8}{*}{4}
& Uncal    & \ding{56} & 13.49 & 23.14 & 15.49 & 17.48 & 20.36 & 20.20 & 51.67 & 43.03 & 51.08 & 25.07 & 19.59 & 15.12 & 47.99 & 20.74 & 15.82 & 26.69 \\
& TS       & \ding{56} & 11.95 & 20.55 & 13.66 & 15.41 & 18.15 & 17.96 & 45.34 & 38.92 & 44.88 & 22.12 & 17.42 & 13.40 & 42.68 & 18.23 & 13.91 & 23.64 \\
& IR       & \ding{56} & 13.08 & 13.20 & 11.59 & 11.95 & 13.42 & 13.87 & 32.78 & 26.30 & 31.84 & 14.46 & 13.05 & 12.41 & 29.90 & 13.27 & 11.99 & 17.54 \\
& AdaTS    & \ding{56} & 11.68 & 21.19 & 13.66 & 14.82 & 17.29 & 18.14 & 47.55 & 39.22 & 45.75 & 21.87 & 17.48 & 13.57 & 44.75 & 17.95 & 13.76 & 23.91 \\
& GC       & \ding{56} &  8.52 & 17.19 & 10.10 & 11.20 & 14.37 & 13.63 & 32.96 & 29.80 & 35.71 & 16.45 & 13.58 &  9.62 & 31.71 & 12.78 & 10.63 & 17.88 \\
& HFE      & \ding{52} &  3.96 &  9.32 &  4.59 &  5.37 &  8.01 &  7.63 & 28.53 & 23.84 & 28.29 & 10.10 &  6.81 &  5.45 & 27.07 &  8.46 &  4.38 & 12.12 \\
& FeatClip & \ding{56} & \underline{2.67} & \underline{6.88} & \underline{3.76} & \underline{3.30} & \underline{5.30} & \underline{5.34} & \underline{17.13} & \underline{15.56} & \underline{15.84} & \underline{6.05} & \underline{5.49} & \underline{3.36} & \underline{16.80} & \underline{4.11} & \underline{3.88} & \underline{7.70} \\
\rowcolor{lightgray}
& \ours    & \ding{56} & \textbf{2.79} & \textbf{4.92} & \textbf{2.95} & \textbf{2.85} & \textbf{4.43} & \textbf{4.92} & \textbf{16.85} & 17.00 & 15.94 & \textbf{5.32} & \textbf{4.13} & \textbf{2.98} & \textbf{16.35} & \textbf{3.76} & \textbf{2.77} & \textbf{7.20} \\
\midrule

\multirow{8}{*}{5}
& Uncal    & \ding{56} & 17.03 & 49.73 & 22.75 & 21.85 & 33.06 & 24.66 & 55.53 & 39.78 & 57.22 & 27.57 & 23.55 & 26.15 & 54.74 & 22.55 & 18.19 & 32.96 \\
& TS       & \ding{56} & 15.04 & 44.99 & 20.11 & 19.23 & 29.74 & 22.04 & 48.71 & 35.87 & 50.28 & 24.35 & 20.86 & 23.25 & 48.53 & 19.94 & 16.04 & 29.27 \\
& IR       & \ding{56} & 12.54 & 31.21 & 13.52 & 12.90 & 19.51 & 15.79 & 35.24 & 23.57 & 37.46 & 14.80 & 14.54 & 15.12 & 35.36 & 14.19 & 12.51 & 20.55 \\
& AdaTS    & \ding{56} & 14.75 & 45.23 & 20.34 & 18.82 & 29.37 & 22.50 & 50.67 & 36.31 & 50.97 & 23.94 & 20.77 & 22.47 & 50.40 & 19.90 & 16.06 & 29.50 \\
& GC       & \ding{56} & 10.85 & 39.89 & 15.74 & 13.51 & 24.77 & 17.21 & 35.87 & 27.40 & 40.47 & 18.25 & 16.46 & 14.97 & 36.42 & 14.73 & 12.49 & 22.60 \\
& HFE      & \ding{52} &  5.37 & 27.36 &  7.52 &  8.20 & 17.51 & 11.17 & 30.62 & 20.48 & 32.03 & 11.19 &  8.81 &  9.16 & 30.89 &  8.87 &  5.12 & 15.62 \\
& FeatClip & \ding{56} & \underline{3.41} & \underline{23.22} & \underline{6.39} & \underline{4.66} & \underline{12.17} & \underline{7.57} & \textbf{18.68} & \textbf{14.04} & \textbf{19.87} & \underline{7.13} & \underline{6.66} & \underline{6.55} & \textbf{19.63} & \underline{5.14} & \underline{4.71} & \underline{10.65} \\
\rowcolor{lightgray}
& \ours    & \ding{56} & \textbf{3.11} & \textbf{19.63} & \textbf{4.51} & \textbf{4.10} & \textbf{10.66} & \textbf{6.64} & \underline{18.74} & \underline{14.87} & \underline{20.29} & \textbf{6.32} & \textbf{5.24} & \textbf{6.51} & \underline{19.67} & \textbf{4.71} & \textbf{3.29} & \textbf{9.89} \\
\bottomrule
\end{tabular}
\end{adjustbox}
\caption{Detailed calibration results on CIFAR-100-C (ECE \%) for ViT-B/16 across all 15 corruption types and five severity levels. $D_{\text{OOD}}$ indicates whether the method requires additional OOD samples (\ding{52}) or not (\ding{56}). Lower is better.}
\label{tab:cifar100c_ece_vitb16_full}
\end{table*}

\section{Extended Results for Medical datasets}

Table~\ref{tab:ece_miccai} reports detailed calibration results across $11$ medical datasets. While the main paper (Table 4) discusses the average performance, this table reveals consistent improvements of \ours{} at the dataset level. In particular, substantial gains are observed on challenging datasets where existing methods struggle. For example, on DermMNIST the ECE for RN-50 drops from $7.44\%$ (GC) and $9.82\%$ (FeatClip) to $\mathbf{2.22\%}$ with \ours{}. Similarly, on OrganCMNIST the ECE reduces from $2.83\%$ (GC) and $4.17\%$ (FeatClip) to $\mathbf{1.75\%}$, and on TissueMNIST from $0.81\%$ (IR) and $2.95\%$ (FeatClip) to $\mathbf{0.57\%}$. 

Consistent improvements are also observed across transformer backbones. For instance, on ViT-S the ECE on BloodMNIST decreases from $0.67\%$ (FeatClip) to $\mathbf{0.39\%}$, while on ViT-B the error on BreastMNIST drops from $5.38\%$ (FeatClip) and $7.78\%$ (IR) to $\mathbf{3.61\%}$. Across several datasets, including OrganSMNIST and PathMNIST, \ours{} also provides stable improvements across multiple backbones. These results confirm that the gains arise from consistent calibration improvements across diverse medical imaging tasks.

\begin{table*}[!h]
\centering
\caption{Expected Calibration Error (ECE) (\%) of calibration methods across $11$ datasets and different backbones.}
\label{tab:ece_miccai}
\resizebox{\linewidth}{!}{%
\begin{tabular}{llccccccccccc|c}
\toprule
\multirow{2}{*}{Backbone} & \multirow{2}{*}{Method} &
\textbf{HAM} & \textbf{Path} & \textbf{Blo} & \textbf{Bre} & \textbf{Derm} &
\textbf{OrgA} & \textbf{OrgC} & \textbf{OrgS} & \textbf{Pneu} & \textbf{Ret} & \textbf{Tiss} & \textbf{Avg} \\
& & \multicolumn{12}{c}{} \\
\midrule

\multirow{7}{*}{RN-50}
& Uncal & 7.78 & 5.41 & 1.66 & 8.81 & 14.59 & 4.17 & 5.04 & 13.69 & 7.45 & 10.25 & 3.93 & 7.53 \\
& TS & 6.95 & 5.40 & 1.44 & 7.75 & 13.12 & 4.07 & 4.49 & 12.22 & 7.24 & 10.24 & 2.60 & 6.87 \\
& IR & 7.40 & 5.33 & \textbf{0.68} & 8.20 & 9.64 & 4.09 & \underline{3.03} & \underline{6.63} & 8.07 & 16.79 & \underline{0.81} & 6.42 \\
& AdaTS & 2.50 & 5.30 & 1.50 & 9.10 & 13.10 & 3.90 & 4.40 & 12.02 & 7.60 & 8.70 & 2.02 & 6.38 \\
& GC & 5.77 & 5.39 & 0.80 & \underline{5.99} & \underline{7.44} & 4.06 & 2.83 & 8.07 & \underline{6.81} & 6.62 & 0.98 & \underline{4.98} \\
& FeatClip & \underline{2.10} & \underline{4.73} & 0.90 & 9.48 & 9.82 & \underline{3.89} & 4.17 & 10.14 & 7.26 & \underline{5.45} & 2.95 & 5.54 \\
& \ours & \textbf{1.84} & \textbf{4.50} & \underline{0.73} & \textbf{6.10} & \textbf{2.22} & \textbf{3.65} & \textbf{1.75} & \textbf{5.51} & \textbf{6.90} & \textbf{5.42} & \textbf{0.57} & \textbf{3.56} \\
\midrule

\multirow{7}{*}{RN-18}
& Uncal & 12.24 & 7.03 & 1.69 & 9.32 & 14.69 & 4.21 & 5.20 & 13.85 & 7.91 & 35.65 & 5.23 & 10.64 \\
& TS & 11.25 & 7.03 & 1.55 & 8.99 & 13.41 & 4.18 & 4.43 & 12.47 & 7.55 & 32.22 & 3.49 & 9.69 \\
& IR & \underline{3.40} & 7.46 & \underline{0.88} & \underline{5.83} & 11.65 & 4.07 & 3.55 & 12.78 & 15.55 & 16.43 & \underline{1.25} & 7.53 \\
& AdaTS & 11.10 & 6.81 & 1.40 & 10.30 & 12.90 & \textbf{3.91} & 4.20 & 11.10 & 7.90 & \underline{8.20} & 1.90 & 7.25 \\
& GC & 10.23 & \underline{7.02} & 1.09 & 7.34 & \underline{8.23} & 4.86 & \underline{2.70} & \underline{8.19} & \underline{6.90} & 8.30 & \underline{0.67} & \underline{5.96} \\
& FeatClip & 11.60 & 6.94 & 1.21 & 8.49 & 8.63 & 4.26 & 3.96 & 10.23 & 7.46 & 8.96 & 1.64 & 6.67 \\
& \ours & \textbf{3.50} & \textbf{6.52} & \textbf{0.74} & \textbf{5.21} & \textbf{2.20} & \underline{4.10} & \textbf{1.54} & \textbf{4.95} & \textbf{6.66} & \textbf{8.10} & \textbf{0.78} & \textbf{4.03} \\
\midrule

\multirow{7}{*}{DN-121}
& Uncal & 11.31 & 5.85 & 1.23 & 8.08 & 14.50 & 4.23 & 4.92 & 12.79 & 8.29 & 36.98 & 2.77 & 10.09 \\
& TS & 9.04 & 5.85 & 1.11 & 7.24 & 13.18 & 3.87 & 4.14 & 11.43 & 7.75 & 34.53 & 1.77 & 9.08 \\
& IR & \underline{3.90} & 6.76 & \textbf{0.53} & \textbf{4.92} & 10.13 & \textbf{2.78} & \underline{3.41} & \underline{4.47} & 7.71 & 33.06 & \underline{0.89} & 7.14 \\
& AdaTS & 8.70 & \underline{5.50} & 1.14 & 8.00 & 12.90 & 3.90 & 3.90 & 11.10 & 9.87 & 12.70 & 1.20 & 7.17 \\
& GC & 7.35 & 5.85 & 0.83 & 6.76 & 7.58 & 3.18 & 2.24 & 6.89 & 7.23 & 11.20 & 0.79 & \underline{5.45} \\
& FeatClip & 4.10 & 5.69 & 0.80 & 9.76 & \underline{7.17} & 3.44 & 3.22 & 7.15 & 7.91 & 8.60 & 2.18 & 5.46 \\
& \ours & \textbf{2.40} & \textbf{5.40} & \underline{0.69} & \underline{5.40} & \textbf{1.83} & \underline{2.82} & \textbf{0.98} & \textbf{3.20} & \textbf{6.97} & \textbf{8.65} & \textbf{0.77} & \textbf{3.56} \\
\midrule

\multirow{7}{*}{ViT-S}
& Uncal & 9.77 & 7.28 & 1.62 & 11.40 & 17.14 & 4.36 & 5.05 & 10.30 & 6.28 & 10.94 & 1.49 & 7.78 \\
& TS & 9.34 & 7.27 & 1.57 & 10.30 & 16.18 & 4.01 & 4.61 & 9.10 & 5.62 & 10.53 & 0.84 & 7.22 \\
& IR & 16.95 & 8.22 & 12.75 & 7.22 & 22.25 & 3.16 & 2.83 & \underline{6.60} & 4.68 & 13.03 & \underline{0.78} & 7.13 \\
& AdaTS & 8.65 & 7.17 & 1.34 & 8.51 & 16.20 & 3.77 & 4.50 & 8.85 & 5.71 & 12.08 & 1.50 & 7.12 \\
& GC & 8.63 & 7.25 & 0.87 & 8.25 & 11.69 & 3.19 & 2.68 & 6.76 & 4.83 & 11.72 & 0.85 & \underline{6.06} \\
& FeatClip & \underline{3.13} & \underline{7.07} & \underline{0.67} & \underline{6.61} & \underline{4.23} & \underline{3.15} & \underline{2.73} & 6.77 & \underline{4.6} & \underline{8.15} & 1.50 & 4.42 \\
& \ours & \textbf{2.79} & \textbf{5.88} & \textbf{0.39} & \textbf{5.11} & \textbf{3.10} & \textbf{2.70} & \textbf{0.78} & \textbf{5.70} & \textbf{4.21} & \textbf{9.30} & \textbf{0.51} & \textbf{3.68} \\
\midrule

\multirow{7}{*}{ViT-B}
& Uncal & 9.88 & 6.15 & 2.15 & 8.95 & 9.27 & 4.63 & 6.32 & 12.10 & 3.82 & 8.61 & 0.46 & 6.58 \\
& TS & 9.11 & 6.14 & 1.74 & 8.05 & 7.24 & 4.08 & 5.76 & 10.60 & \underline{4.02} & 7.85 & \underline{0.45} & 5.91 \\
& IR & 9.28 & 6.07 & \underline{0.81} & \underline{7.78} & \underline{3.31} & \underline{3.02} & 10.21 & \underline{6.21} & 5.64 & 9.82 & 0.91 & 5.73 \\
& AdaTS & 8.70 & \underline{5.93} & 1.80 & 8.98 & 6.44 & 3.85 & 5.80 & 8.14 & 5.86 & \textbf{7.20} & 0.73 & 5.97 \\
& GC & 9.12 & 6.11 & 1.10 & 6.41 & 2.73 & 2.97 & 3.71 & 9.57 & 5.80 & 6.40 & 0.59 & \underline{4.96} \\
& FeatClip & \underline{2.80} & \underline{5.13} & 0.86 & \underline{5.38} & \underline{2.28} & \underline{2.81} & \underline{2.83} & 7.64 & \textbf{3.78} & 8.82 & 0.62 & 3.91 \\
& \ours & \textbf{2.26} & \textbf{4.87} & \textbf{0.77} & \textbf{3.61} & \textbf{1.78} & \textbf{2.32} & \textbf{1.86} & \textbf{5.50} & 4.43 & \underline{7.64} & \textbf{0.44} & \textbf{3.23} \\
\bottomrule
\end{tabular}%
}
\end{table*}

\section{AECE in standard datasets}

\begin{table}[!h]
\centering
\scriptsize
\caption{\textbf{Calibration performance (AECE$\downarrow$) in standard image classification benchmarks.} 
Results across datasets and backbones. Best method in \textbf{bold}, second best \underline{underlined}. 
Performance differences are computed with respect to the strongest competing baseline per row.}
\label{tab:main_clean_adaece}
\begin{tabular}
{p{1.6cm}lP{1cm}P{1.cm}P{1.cm}P{1.cm}P{1.cm}P{1.3cm}P{1.3cm}}

\toprule
\textbf{Dataset} & \textbf{Backbone} & \textbf{Uncal} & \textbf{TS}& \textbf{IR} &  \textbf{AdaTS} & \textbf{GC} &\textbf{FeatClip} & \ours \\
\midrule

\multirow{5}{*}{CIFAR-10}
 & RN-50  & 1.70 & 1.30 & \underline{0.43} & 0.76 & 0.86 & 0.57 & \textbf{0.19}\imp{-0.32} \\
 & RN-18  & 1.96 & 1.32 & \underline{0.59} & 0.70 & 1.33 & 0.78 & \textbf{0.44}\imp{-0.15} \\
 & DN-121 & 1.44 & 1.29 & 0.72 & 0.50 & 0.96 & \underline{0.40} & \textbf{0.15}\imp{-0.14} \\
 & ViT-S  & 1.80 & 1.51 & \underline{0.45} & 1.01 & 1.42 & 0.68 & \textbf{0.42}\imp{-0.03} \\
 & ViT-B  & 1.97 & 1.49 & 0.69 & 1.16 & 1.19 & \underline{0.65} & \textbf{0.35}\imp{-0.30} \\
\midrule

\multirow{5}{*}{CIFAR-100}
 & RN-50  & 8.62 & 5.90 & 3.08 & 3.43 & 3.99 & \underline{2.90} & \textbf{2.86}\imp{-0.04} \\
 & RN-18  & 7.87 & 4.22 & 1.78 & 2.40 & 2.92 & \underline{1.53} & \textbf{1.32}\imp{-0.21} \\
 & DN-121 & 7.60 & 4.83 & 2.91 & 2.58 & 3.12 & \underline{2.51} & \textbf{2.44}\imp{-0.14} \\
 & ViT-S  & 7.98 & 4.31 & 2.73 & 2.77 & 3.64 & \underline{2.80} & \textbf{2.56}\imp{-0.24} \\
 & ViT-B  & 10.79 & 6.89 & 4.67 & 4.48 & 5.34 & \underline{2.39} & \textbf{2.26}\imp{-0.13} \\
\midrule

\multirow{5}{*}{ImageNet}
 & RN-50  & 3.71 & 2.26 & 1.97 & 2.06 & 2.39 & \underline{1.80} & \textbf{1.68}\imp{-0.12} \\
 & RN-18  & 2.61 & 1.86 & 1.56 & 1.79 & 1.67 & \underline{1.49} & \textbf{1.34}\imp{-0.15} \\
 & DN-121 & 2.61 & 1.73 & 1.75 & 1.78 & 1.85 & \underline{1.35} & \textbf{1.31}\imp{-0.04} \\
 & ViT-S  & 2.01 & 1.51 & 1.11 & 1.25 & 1.57 & \underline{1.48} & \textbf{1.03}\imp{-0.08} \\
 & ViT-B  & 5.77 & 3.28 & 2.87 & 2.88 & 3.17 & \underline{2.72} & \textbf{2.59}\imp{-0.13} \\
\bottomrule
\end{tabular}
\end{table}

In addition to ECE results of the main paper, we also report the Adaptive Expected Calibration Error (AECE), which uses equal-mass binning to better capture calibration behavior across the confidence spectrum. Specifically, predictions are first sorted by confidence and partitioned into $M$ bins such that each bin contains approximately $N/M$ samples. The AECE is then computed as:
$
\frac{1}{M} \sum_{m=1}^{M}
\left|
\operatorname{acc}(b_m) - \operatorname{conf}(b_m)
\right|,
$
where $\operatorname{acc}(b_m)$ and $\operatorname{conf}(b_m)$ denote the average accuracy and confidence within bin $b_m$. Unlike ECE, which uses fixed-width bins and weights errors by bin frequency, AECE ensures that each bin contains a comparable number of samples, making the metric more sensitive to calibration errors across the entire confidence distribution.

Table~\ref{tab:main_clean_adaece} reports AECE results on CIFAR-10, CIFAR-100 and ImageNet across five architectures. Overall, \ours{} achieves the lowest AECE in all settings, consistently outperforming existing post-hoc calibration methods. On CIFAR-10, \ours{} reduces AECE to $0.15$ on DN-121 and $0.19$ on RN-50, improving upon the strongest baseline by margins up to $0.32$. On CIFAR-100, where calibration errors are substantially larger, \ours{} again provides the best performance across all backbones, achieving $1.32$ on RN-18 and $2.26$ on ViT-B. Similar trends hold on ImageNet, where \ours{} attains the lowest AECE across architectures, including $1.03$ on ViT-S and $1.31$ on DN-121. These results demonstrate that \ours{} consistently improves calibration under the evaluation of AECE.

\section{AECE in imbalanced datasets}

\begin{table}[!h]
\centering
\scriptsize
\caption{\textbf{Calibration performance (AECE$\downarrow$) under long-tailed distributions.} 
Results on CIFAR-100-LT (R-10, R-100) and ImageNet-LT. 
Best method in \textbf{bold}, second best \underline{underlined}. 
Performance differences are computed with respect to the strongest competing baseline per row.}
\label{tab:imbalance_adaece}
\begin{tabular}
{p{1.8cm}lP{1cm}P{1.cm}P{1.cm}P{1.cm}P{1.cm}P{1.2cm}P{1.2cm}}
\toprule
\textbf{Dataset} & \textbf{Backbone} & \textbf{Uncal} & \textbf{TS}& \textbf{IR} & \textbf{AdaTS} & \textbf{GC} & \textbf{FeatClip} & \ours \\
\midrule

\multirow{5}{*}{R-10(mild)}
 & RN-50  & 19.53 & 11.21 & \underline{5.37} & 6.01 & 8.05 & 9.48 & \textbf{3.76}\imp{-1.61} \\
 & RN-18  & 19.11 & 8.69 & 5.31 & \underline{4.48} & 6.78 & 4.73 & \textbf{3.54}\imp{-0.94} \\
 & DN-121 & 19.13 & 9.16 & \underline{5.44} & 5.46 & 7.38 & 6.63 & \textbf{3.66}\imp{-1.78} \\
 & ViT-S  & 19.14 & 10.05 & 5.88 & 6.68 & 8.36 & \underline{3.55} & \textbf{3.46}\imp{-0.09} \\
 & ViT-B  & 24.12 & 14.30 & 15.24 & 12.48 & 12.60 & \underline{3.63} & \textbf{3.50}\imp{-0.13} \\
\midrule

\multirow{5}{*}{R-100(extreme)}
 & RN-50  & 48.64 & 28.81 & 27.24 & 23.54 & 25.84 & \underline{5.82} & \textbf{5.06}\imp{-0.76} \\
 & RN-18  & 45.81 & 21.73 & 25.89 & 29.73 & 21.51 & \underline{2.98} & \textbf{2.93}\imp{-0.05} \\
 & DN-121 & 47.27 & 24.12 & 26.86 & 21.94 & 23.63 & \underline{4.99} & \textbf{3.07}\imp{-1.92} \\
 & ViT-S  & 50.00 & 31.42 & 30.04 & 26.12 & 29.74 & \underline{9.73} & \textbf{8.19}\imp{-1.54} \\
 & ViT-B  & 52.51 & 34.79 & 33.82 & 28.36 & 32.89 & \underline{8.43} & \textbf{8.37}\imp{-0.06} \\
\midrule

\multirow{5}{*}{ImageNet-LT}
 & RN-50  & 3.69 & 4.97 & 5.10 & 3.88 & 4.77 & \underline{3.32} & \textbf{2.17}\imp{-1.15} \\
 & RN-18  & 2.63 & 2.97 & 2.63 & 2.54 & 4.56 & \underline{2.36} & \textbf{2.22}\imp{-0.14} \\
 & DN-121 & 2.51 & 2.50 & 2.94 & 2.74 & 2.66 & \underline{1.66} & \textbf{1.12}\imp{-0.54} \\
 & ViT-S  & 1.97 & 3.80 & 2.71 & 1.99 & 2.64 & \bf{1.87} & \underline{1.95}\imp{+0.08} \\
 & ViT-B  & 5.60 & 5.27 & 5.63 & 5.85 & 5.63 & \underline{5.59} & \textbf{4.15}\imp{-1.44} \\
\bottomrule
\end{tabular}
\end{table}

We also report the AECE results on long-tailed benchmarks (refer Table~\ref{tab:imbalance_adaece}) including CIFAR-100-LT (R-10, R-100) and ImageNet-LT. Compared to ECE, AECE is particularly informative in long-tailed settings because equal-mass binning prevents the metric from being dominated by frequent head-class predictions and ensures that calibration errors in low-confidence or tail regions are adequately reflected.

Across all imbalance regimes, \ours{} consistently achieves the lowest AECE in nearly all settings. Under mild imbalance (R-10), \ours{} reduces AECE substantially across architectures, for example improving from $5.37$ to $\mathbf{3.76}$ on RN-50 and from $5.44$ to $\mathbf{3.66}$ on DN-121. Similar trends hold for transformer backbones, where \ours{} achieves $\mathbf{3.46}$ on ViT-S and $\mathbf{3.50}$ on ViT-B. 
Under extreme imbalance (R-100), the improvements remain pronounced despite the much larger calibration errors. For instance, AECE drops from $27.24$ (IR) to $\mathbf{5.06}$ on RN-50 and from $4.99$ to $\mathbf{3.07}$ on DN-121. On ViT-S, \ours{} reduces AECE from $9.73$ to $\mathbf{8.19}$. Finally, on ImageNet-LT, \ours{} again achieves the best performance across most backbones, including $2.17$ on RN-50 and $1.12$ on DN-121, indicating improved calibration even in large-scale long-tailed recognition scenarios. Overall, these results demonstrate that \ours{} effectively mitigates calibration errors under class imbalance, particularly in regions emphasized by the AECE metric.

\section{Reliability analysis}

\begin{figure*}[!h]
\centering
\includegraphics[width=\textwidth]{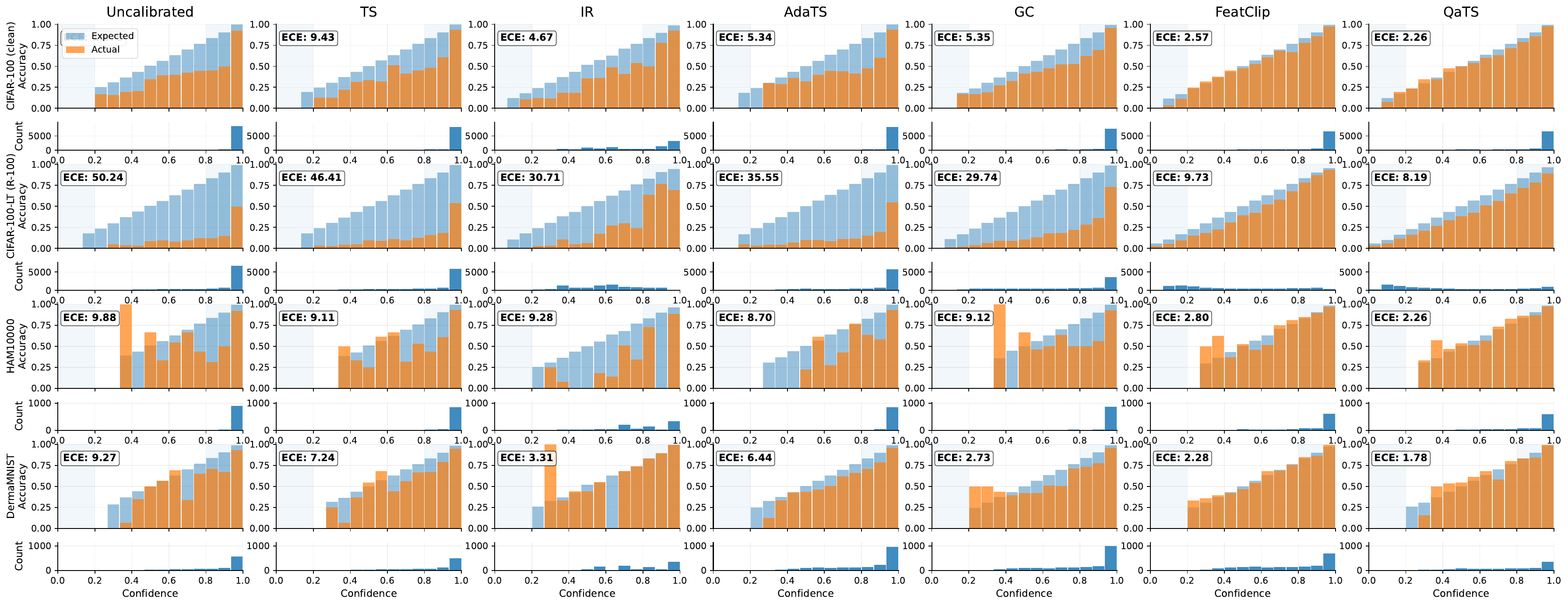}
\caption{
\textbf{Reliability diagrams across standard, long-tailed, and medical benchmarks.}
Comparison of post-hoc calibration methods on CIFAR-100 (clean), CIFAR-100-LT (R-100), HAM10000, and DermaMNIST datasets. 
For each method, the top plot shows expected confidence ({\color{blue}blue}) and empirical accuracy ({\color{orange}orange}) across confidence bins, while the bottom histogram shows the number of samples per bin. Best viewed in color.
}
\label{fig:rel_dig}
\end{figure*}
In addition to the numerical ECE and AECE results, we also visualize calibration behavior through reliability diagrams in Figure~\ref{fig:rel_dig} for standard, long-tailed, and medical benchmarks, comparing \ours{} against existing post-hoc baselines. Across all settings, the uncalibrated model exhibits a clear mismatch between empirical accuracy and confidence, with the discrepancy becoming particularly severe under long-tailed imbalance and in medical datasets. Classical calibration methods such as TS, IR, AdaTS, and GC partially reduce this gap, but still leave noticeable deviations across several confidence regions.

In contrast, \ours{} yields reliability curves which align expected confidence with observed accuracy more closely  across the bins. On CIFAR-100 (clean), \ours{} substantially improves calibration over the standard baselines and remains competitive with the strongest recent methods, while preserving predictive accuracy. On CIFAR-100-LT (R-100), shown here with ViT-S/16, the benefit of \ours{} is even clearer: the gap between confidence and accuracy is consistently reduced across bins, indicating that the proposed quantile-adaptive scaling is particularly effective under the heterogeneous confidence distributions induced by long-tailed data. Similar behavior is observed on HAM10000 and DermaMNIST, where \ours{} produces smoother and better-aligned reliability plots, especially in the medium- and high-confidence regions.
A further observation from these plots is that some competing methods, particularly FeatClip, can achieve competitive calibration on selected settings, but often do so by modifying the prediction itself, which comes at the cost of reduced classification accuracy (refer to Section 5.1 and Fig. 2 of the main paper). In contrast, \ours{} improves calibration while preserving the original accuracy.


\section{Piecewise linear function}
\label{sec:piecewise}
We define here the piecewise monotone linear function employed in the last experiment of the main manuscript (Section 5.F). Let $q_1<q_2<\dots<q_{K+1}$ denote the quantile boundaries, and $t_i$ the learnable temperature value at each knot, for any prediction with quantile $q(x) \in [q,q_{i+1})$, the temperature value is obtained by linear interpolation: 
\begin{equation}
    T(q)=t_i+\frac{q-q_i}{q_{i+1}-q_i}(t_{i+1}-t_j).
\end{equation}
This parameterization strictly generalizes the linear model while preserving monotonicity, and allows the temperature to vary more flexibly across the quantile axis, containing $K+1$ learnable parameters.




\end{document}